\documentclass{article} 
\PassOptionsToPackage{table}{xcolor}
\usepackage{iclr2027_conference,times}

\usepackage{amsmath,amsfonts,bm}

\def\eqref#1{equation~\ref{#1}}

\def\1{\bm{1}}

\DeclareMathAlphabet{\mathsfit}{\encodingdefault}{\sfdefault}{m}{sl}
\SetMathAlphabet{\mathsfit}{bold}{\encodingdefault}{\sfdefault}{bx}{n}

\usepackage{hyperref}
\usepackage{url}

\definecolor{citationgreen}{HTML}{285C83}
\definecolor{referenceblue}{HTML}{397F88}
\hypersetup{
  colorlinks=true,
  citecolor=citationgreen,
  linkcolor=referenceblue,
  urlcolor=referenceblue
}

\DeclareRobustCommand{\microclipicon}{%
  \raisebox{-0.15em}{\includegraphics[height=1em]{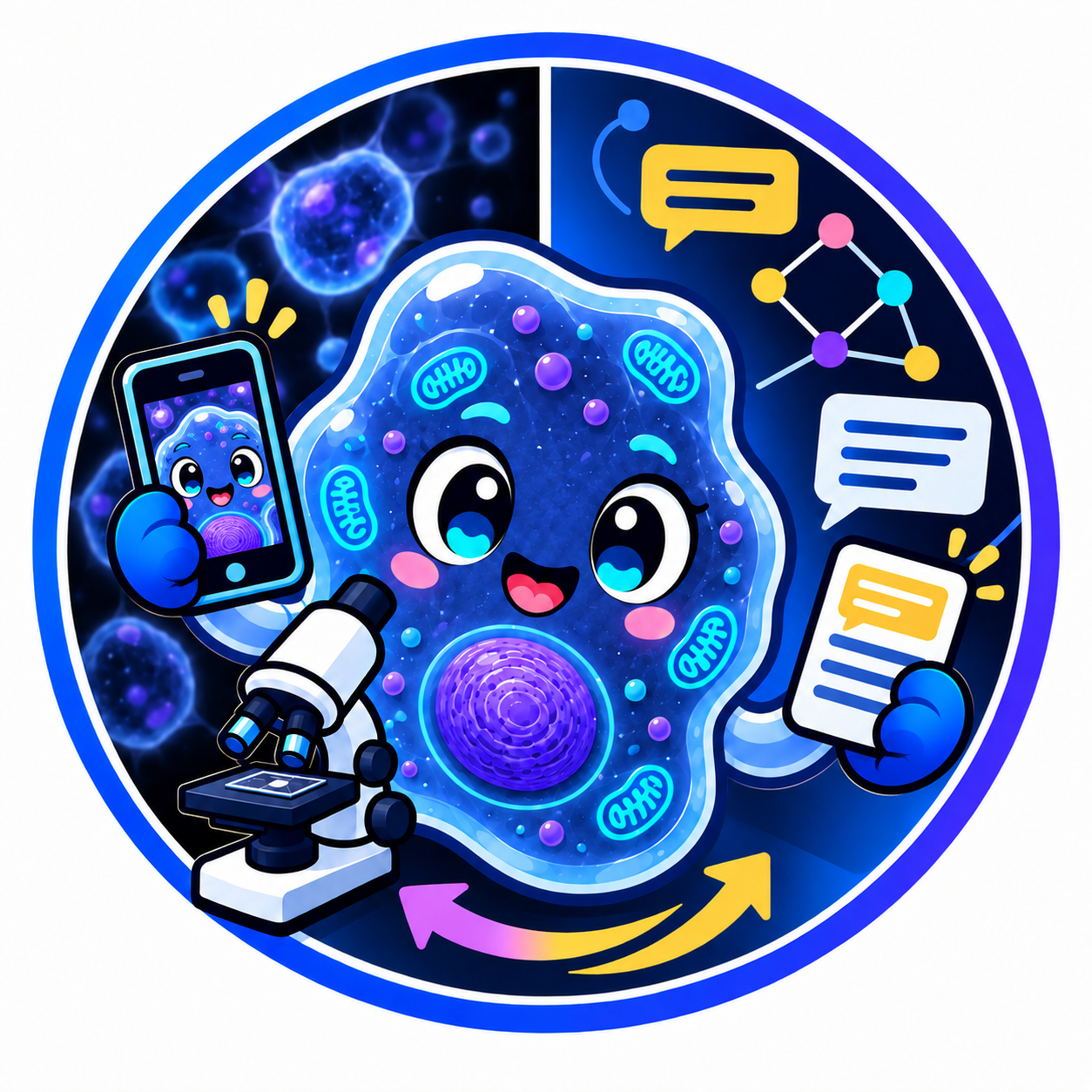}}%
  \hspace{0.25em}%
}
\title{\microclipicon{}MicroCLIP: Connecting Text and Microscopy Images via Contrastive Language--Image Pretraining}

\author{%
\parbox{\textwidth}{\centering
\textbf{Yanran Ruan}\textsuperscript{1*}, \textbf{Zhixiang Wei}\textsuperscript{1*},
\textbf{Shiyao Hong}\textsuperscript{1}, \textbf{Xiaoxiao Ma}\textsuperscript{1},\\
\textbf{Yuxuan Gu}\textsuperscript{1}, \textbf{Liming Ye}\textsuperscript{1},
\textbf{Huaian Chen}\textsuperscript{1\ensuremath{\dagger}}, \textbf{Yi Jin}\textsuperscript{1\ensuremath{\dagger}}\\[0.2em]
\normalfont
{\small\textsuperscript{1}University of Science and Technology of China}\\[0.2em]
}
}

\iclrfinalcopy
\begin{document}

\maketitle

\renewcommand{\thefootnote}{}
\footnotetext{\textsuperscript{*}Equal contribution.\quad
\textsuperscript{\ensuremath{\dagger}}Corresponding authors.}

\begin{figure}[h]
\vspace{-1pt} 
\centering
\includegraphics[width=0.95\textwidth]{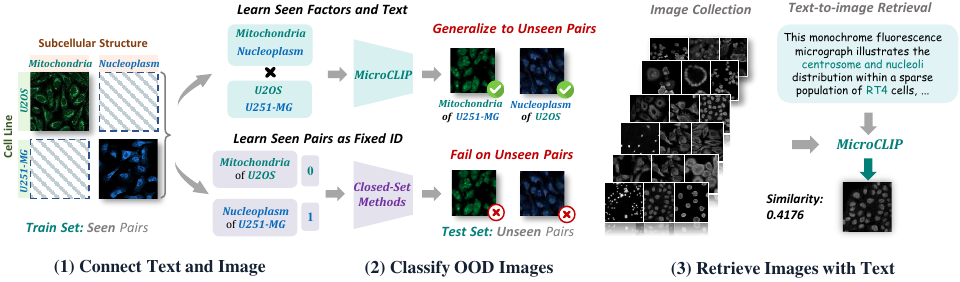}
\caption{
(1) We connect fluorescence microscopy images with text, so that subcellular structures and cell lines can both be represented in a shared semantic space rather than as isolated class IDs.
(2) Building on this shared semantic space, MicroCLIP generalizes to unseen structure--cell-line pairs, whereas conventional closed-set methods are restricted to seen pairs.
(3) Beyond recognition, this shared representation further enables language-based retrieval, allowing users to search microscopy images through natural language descriptions.
}
\label{fig:teaser}
\end{figure}

\vspace{8pt} 

\begin{abstract}
Understanding fluorescence microscopy images is fundamental to automating biological research. However, existing methods typically learn predefined concepts from closed-set data, limiting transfer across tasks and biological contexts. In this work, we introduce MicroCLIP, a vision–language model that learns generalizable cellular and subcellular representations from natural language supervision for open-set image classification and retrieval. To pretrain MicroCLIP, we construct MicroPool-2M, a dataset of approximately 2.5 million fluorescence microscopy image–text pairs spanning diverse cellular structures, cell lines, and imaging conditions. Aligning microscopy images with biological concepts in natural language is challenging because most images lack descriptive captions and biological knowledge annotations. We therefore progressively establish image–text associations at three levels to provide MicroCLIP with richer semantic supervision: (1) category-level alignment links images to standardized labels defined by subcellular structure–cell-line combinations; (2) knowledge-level alignment incorporates biological priors derived from these labels to capture relationships across species and cell types; and (3) detail-level alignment uses a multimodal large language model to recaption and verify image–text pairs at scale, emphasizing fine-grained visual characteristics. Extensive experiments demonstrate that MicroCLIP achieves leading performance in zero-shot classification and text–image retrieval on microscopy images. It retrieves images with specific cellular characteristics from thousands of candidates and generalizes to unseen biological concepts without additional supervision. These results highlight MicroCLIP’s potential as a foundation model for fluorescence microscopy. Models, training code, and datasets are available at \url{https://github.com/Y2R-mv1p/MicroCLIP}.

\end{abstract}

\section{Introduction}

Large-scale fluorescence microscopy generates massive amounts of biological imaging data, creating a growing need for automated image analysis \citep{ouyang2019analysis,caicedo2017data}. A fundamental requirement for such automation is the reliable recognition of biological content in microscopy images. In protein-localization microscopy, images reveal the spatial distribution of proteins across subcellular structures \citep{thul2017subcellular}. The fluorescence pattern associated with a given subcellular structure can vary across cell lines, making both subcellular structure and cell line relevant factors for recognition \citep{zhang2025pups}. Yet available datasets cover only a subset of their possible combinations, leaving many biological contexts unobserved during training. Models intended for broad use must therefore generalize beyond the structures, cell lines, and structure--cell-line combinations observed during training. This is difficult for conventional closed-set classifiers, whose predictions are restricted to a predefined label space \citep{bendale2016towards}.


Language provides a natural interface for moving beyond this fixed label space, allowing recognition targets to be specified through text rather than tied to class IDs alone \citep{radford2021learning}. For fluorescence microscopy, however, realizing this potential requires grounding biological semantics in image content. On the semantic side, category annotations identify subcellular structures and cell lines, while biological knowledge describes their properties and functions. On the visual side, microscopy images exhibit instance-specific fluorescence patterns and cellular morphology. Establishing a reliable correspondence between these semantics and visual patterns remains challenging. As illustrated in Section~\ref{sec:observation}, a model can describe relevant biological knowledge and image appearance yet fail to identify the correct biological category. This gap between biological semantics and microscopy images limits reliable language-guided recognition.

Existing approaches use visual information and biological semantics in different ways, but remain limited for language-guided microscopy recognition. Image-only methods such as cytoself \citep{kobayashi2022self} and Cell-DINO \citep{moutakanni2025celldino} learn microscopy representations without language supervision. General-purpose vision--language models such as CLIP \citep{radford2021learning} and SigLIP 2 \citep{tschannen2025siglip} align images and text broadly, but lack microscopy-specific biological supervision. Domain-specific multimodal methods move closer, yet target specialized tasks: FluoCLIP \citep{park2026fluoclip} uses stain semantics for focus quality assessment, while CellCLIP \citep{lu2025cellclip} relates cellular phenotypes to perturbation descriptions.



To capture the correspondence between microscopy images and biological semantics, we curate \textbf{MicroPool-2M}, the largest fluorescence microscopy image--text dataset to date, comprising 2.5 million single-channel images paired with textual descriptions. For each image, MicroPool-2M provides natural language descriptions at three levels: category-level captions specify the subcellular structure and cell line, knowledge-level captions describe their biological properties, and detail-level captions capture the fluorescence patterns visible in the image. Together, these image--text pairs cover diverse cell types and subcellular structures, as well as a broad range of biological knowledge.

Building on MicroPool-2M, we introduce \textbf{MicroCLIP}, a vision--language model for fluorescence microscopy. Through contrastive learning, MicroCLIP aligns microscopy images with natural language descriptions at three levels, learning representations that generalize to unseen combinations of cell lines and subcellular structures. It enables biological researchers to use natural language to search large-scale microscopy image collections for cellular morphologies and subcellular structures of interest, without task-specific fine-tuning. As shown in Table~\ref{tab:classification_retrieval}, MicroCLIP outperforms the evaluated general-purpose vision--language models in zero-shot classification and bidirectional image--text retrieval. On unseen structure--cell-line combinations, it achieves 67.70\% classification accuracy, compared with 5.12\% for the strongest general-purpose baseline.

Our contributions are threefold:
\begin{itemize}
\item We identify a biological grounding gap in fluorescence microscopy, highlighting the need to explicitly connect microscopy appearance with biological semantics for language-guided image understanding.


\item We introduce \textbf{MicroPool-2M}, a large-scale multi-source fluorescence image--text dataset that grounds biological semantics in microscopy images through category-level, knowledge-level, and detail-level captions.

\item We develop \textbf{MicroCLIP}, which learns a shared microscopy image--text representation from this multi-level supervision, substantially improving zero-shot generalization while naturally extending fluorescence microscopy analysis to bidirectional language-based retrieval.

\end{itemize}

\section{Related Work}
\paragraph{Representation Learning for Cellular Microscopy.}
Self-supervised methods such as cytoself~\citep{kobayashi2022self} and Cell-DINO~\citep{moutakanni2025celldino} learn microscopy representations for protein localization and cellular characterization without language supervision. PUPS~\citep{zhang2025pups} incorporates protein sequences and cellular landmarks to predict localization in unseen proteins and cell lines. Cross-modal methods connect cellular images to other biological information: CLOOME~\citep{sanchezfernandez2023cloome} aligns images with chemical structures, while CellCLIP~\citep{lu2025cellclip} aligns Cell Painting phenotypes with textual descriptions of perturbations. For brain histology, \citet{sutton2026cytoarchitecture} combine label-linked literature with image-specific properties to supervise cytoarchitectural description generation. These approaches establish complementary links between microscopy and biological information. MicroCLIP focuses on language-guided recognition and retrieval of subcellular structures across cell lines through a shared fluorescence image--text representation.

\paragraph{Vision--Language Pretraining and Supervision Construction.}
CLIP~\citep{radford2021learning} establishes image--text pretraining as a framework for zero-shot recognition and retrieval. SigLIP~\citep{zhai2023sigmoid}, SigLIP~2~\citep{tschannen2025siglip}, and PE-Core~\citep{bolya2025perception} advance general-purpose representations through improved objectives and training recipes. Data construction is another central direction: MetaCLIP~\citep{xu2024demystifying} balances metadata distributions, and DFN~\citep{fang2024data} learns to filter image--text data. On the text side, LaCLIP~\citep{fan2023improving} samples original captions and language rewrites, while VeCLIP~\citep{lai2024veclip} enriches captions with visual information and mixes them with original text. These studies motivate attention to both supervision content and sampling. MicroCLIP applies this perspective to fluorescence microscopy, pairing images with complementary category identities, biological knowledge, and instance-specific visual descriptions while retaining the standard contrastive objective.

\paragraph{Microscopy Resources and Generalization Evaluation.}
The Human Protein Atlas~\citep{thul2017subcellular} and OpenCell~\citep{cho2022opencell} provide annotated protein-localization images, while RxRx1~\citep{sypetkowski2023rxrx1} and Cell Painting resources~\citep{chandrasekaran2024jump} capture variation across experimental batches and biological perturbations. MicroPool-2M integrates multiple microscopy sources into a common annotation vocabulary and augments images with three levels of text supervision. Its evaluation distinguishes unseen structures, unseen cell lines, and unseen pairings, complemented by an external source-shift benchmark. The unseen-pair setting relates to compositional zero-shot learning, which evaluates novel combinations of familiar concepts~\citep{nagarajan2018operators,nayak2023csp}; unseen-factor evaluation is also studied in open-vocabulary compositional learning~\citep{saini2024ovczsl}. We use these distinctions to assess the transfer of microscopy image--text representations alongside language-based retrieval.

\section{Observation: The Biological Grounding Gap}
\label{sec:observation}

\begin{figure}[h]
\centering
\includegraphics[width=0.95\textwidth]{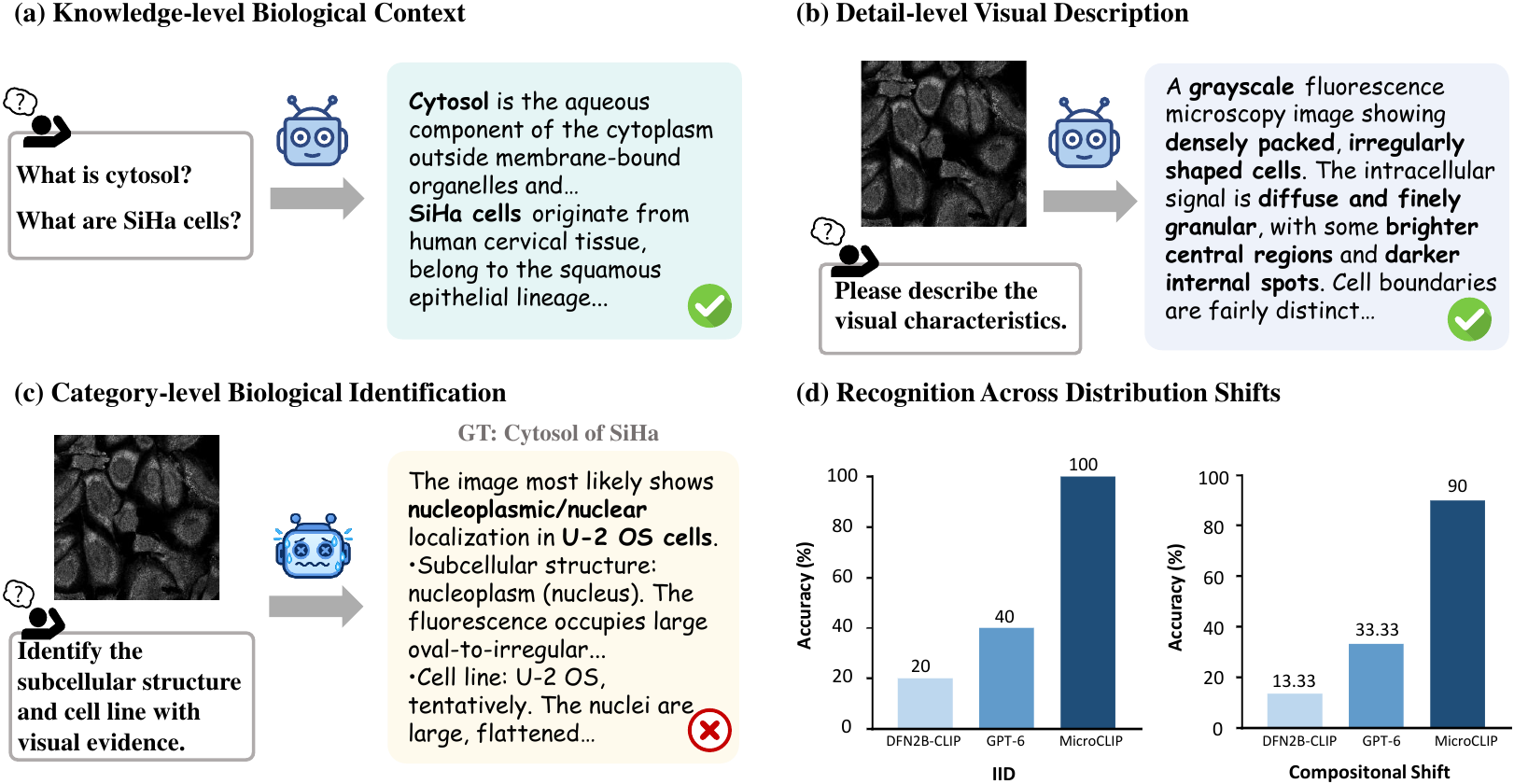}
\caption{
Illustration of the biological grounding gap.
(a--c) GPT-6 responses to biological knowledge, visual description,
and biological identification queries, respectively.
The predicted identity in (c) differs from the reference annotation.
(d) Joint structure--cell-line classification accuracy of DFN2B-CLIP,
GPT-6, and MicroCLIP on diagnostic subsets of IID and Compositional
Shift, with 10 candidate categories and 30 images per split.
}
\label{fig:observation}
\end{figure}

Figure~\ref{fig:observation} illustrates the biological grounding gap
using GPT-6. In response to text queries, the model describes the
biological properties of cytosol and SiHa cells. Given a microscopy
image, it describes diffuse, granular fluorescence and cellular
morphology. Yet when asked to identify the subcellular structure
and cell line, it predicts nucleoplasm in U-2 OS cells rather than
the annotated cytosol in SiHa cells. This example shows that the
ability to describe biological knowledge and visual patterns does
not guarantee correct biological identification. We refer to this
difficulty in reliably associating microscopy images with their
biological semantics as the biological grounding gap.

To examine this difficulty beyond a single example, we select
10 categories from each of the IID and Compositional Shift subsets,
with three images per category. Each image is classified among
the 10 structure--cell-line candidates selected for its split.
As shown in Figure~\ref{fig:observation}(d), GPT-6 achieves joint
category accuracies of 40.00\% (12/30) on IID and 33.33\% (10/30)
on Compositional Shift, whereas MicroCLIP achieves 100.00\% (30/30)
and 90.00\% (27/30), respectively. These results show that GPT-6's
biological identification remains unreliable in this diagnostic
even when candidate identities are supplied, while MicroCLIP
achieves higher accuracy on the same subsets. This small evaluation
does not establish the cause of the errors or imply that every
cell-line identity is uniquely recoverable from a single fluorescence
image. Sampling, prompts, and full comparison results are provided
in Appendix~\ref{app:grounding_diagnostic}. These observations motivate learning a more reliable correspondence
between microscopy images and biological semantics.

\section{MicroPool-2M: A Large-Scale Microscopy Image--Text Dataset}
\label{sec:micropool}

Motivated by the biological grounding gap identified in
Section~\ref{sec:observation}, we construct MicroPool-2M to provide
explicit supervision linking microscopy images to biological semantics.
Each fluorescence microscopy image is paired with three complementary
levels of text: category-level captions specify its annotated biological
identity, knowledge-level captions provide associated biological context,
and detail-level captions describe its visible appearance.
Together, these captions provide the supervision used to train MicroCLIP
to associate biological meaning with microscopy appearance.

\paragraph{Image collection and annotation harmonization.}

Covering diverse biological contexts requires varied image sources and
consistent category definitions. We aggregate single-channel fluorescence images from the Human Protein Atlas~\citep{thul2017subcellular}, Image Data Resource~\citep{williams2017image}, RxRx~\citep{sypetkowski2023rxrx1}, Broad Bioimage Benchmark Collection~\citep{ljosa2012annotated}, and curated literature. Source-specific aliases are mapped to a common vocabulary, assigning each image $x_i$ a structure label $s_i$ and cell-line identity $c_i$. A structure label can denote a single structure or marker, or a combination of them. Each category is represented as $e_i=(s_i,c_i)$. We remove content-identical images before constructing training and evaluation sets.

\paragraph{Three-level text supervision.}
\label{sec:three_level_associations}

Category labels identify biological entities but do not describe their
biological context or the visual appearance of individual images.
We therefore pair each image $x_i$ with category-level, knowledge-level,
and detail-level text:
\begin{equation}
t_i^{\mathrm{cat}} = \mathcal{T}(s_i,c_i), \quad
t_i^{\mathrm{know}} = \operatorname{Concat}\!\left(\mathcal{K}_{s}(s_i),\,\mathcal{K}_{c}(c_i)\right), \quad
t_i^{\mathrm{det}} = \mathcal{D}(x_i,e_i).
\label{eq:three_level_text}
\end{equation}

Here, $\mathcal{T}$ applies the category template:
\text{\textit{a microscopy photo of} \{\textit{structure}\}\_of\_\{\textit{cell}\}}.
The mappings $\mathcal{K}_{s}$ and $\mathcal{K}_{c}$ retrieve all available
biological descriptions generated by GPT-5.6 Sol for the associated
structures or markers and cell line, respectively; composite structure
labels include descriptions of all constituent structures.
For $\mathcal{D}$, we prompt Qwen3.5-397B-A17B with the image and the
normalized category name corresponding to $e_i$ to describe fluorescence
distribution, morphology, signal intensity, and background.
Appendix~\ref{app:caption_construction} provides prompts and caption examples.

\paragraph{Dataset coverage.}


MicroPool-2M contains 3,301,672 images and 3,391 observed structure--cell-line pairs. After data cleaning and split construction, the training split contains 2,505,017 images spanning 2,032 structure--cell-line categories. The resulting annotations, knowledge tables, and image descriptions define the supervision used by MicroCLIP; the evaluation subsets and protocols are specified in Section~\ref{sec:experimental_settings}.

\section{MicroCLIP: Grounding Biological Semantics in Microscopy}
\label{sec:microclip}

\begin{figure}[t]
\centering
\includegraphics[width=0.95\textwidth]{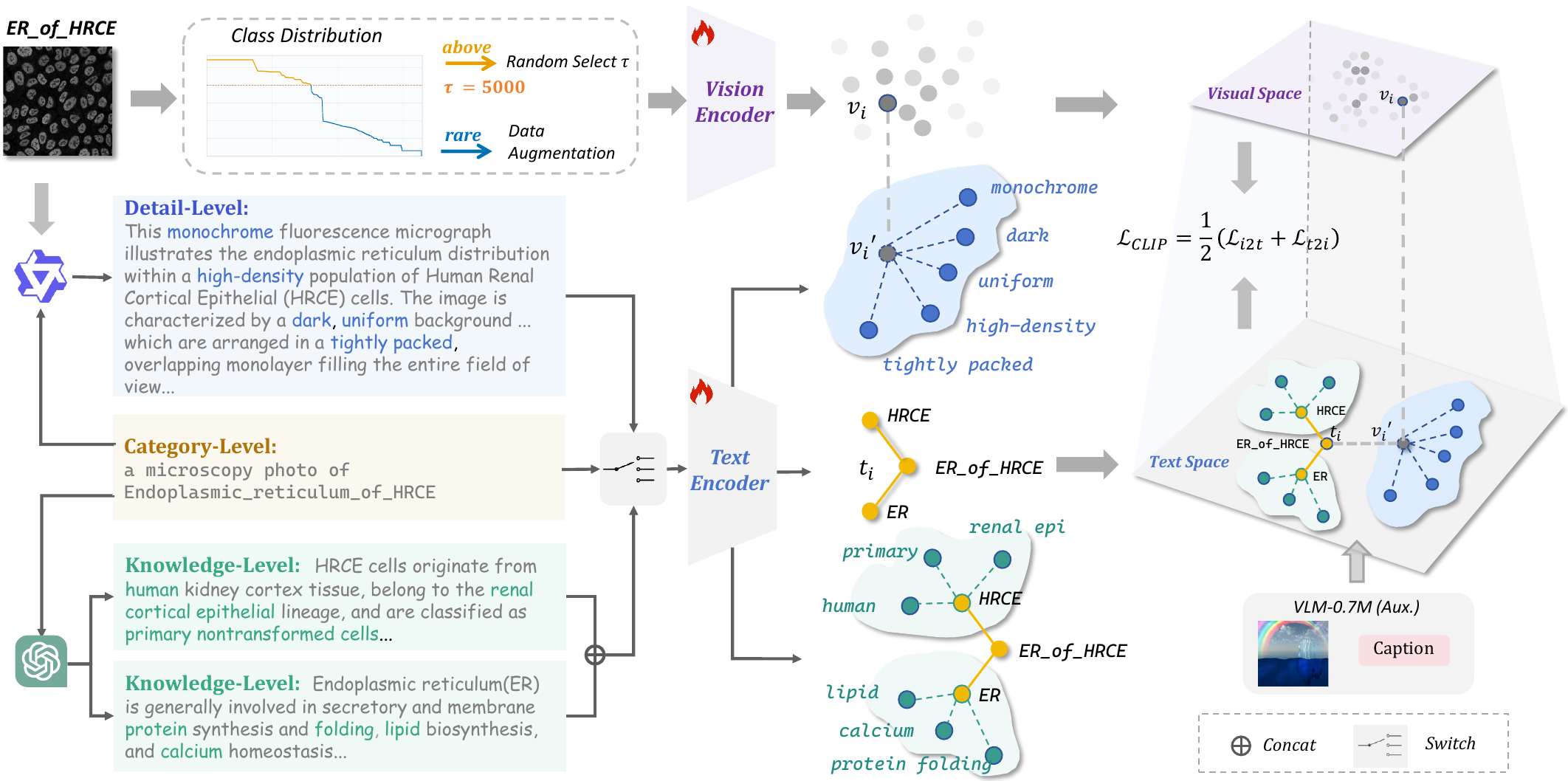}
\caption{MicroCLIP learns a shared image--text representation from category-level, knowledge-level and detail-level supervision. Each microscopy image is paired with one sampled caption for CLIP training. Composition balancing and rare-class augmentation adjust training exposure, while general-domain image--text pairs provide auxiliary supervision.}
\label{fig:framework}
\end{figure}

Building on MicroPool-2M, MicroCLIP learns to associate microscopy
appearance with biological semantics in a shared image--text space.
The three caption levels connect each image to its categorical identity,
associated biological knowledge, and instance-specific visual traits.
We use the CLIP dual-encoder architecture and symmetric contrastive
objective, sampling one caption level for each image occurrence during
training so that repeated observations expose the model to complementary
textual descriptions. Composition-aware sampling balances training
exposure across structure--cell-line pairs, while auxiliary general-domain
image--text data provide broader visual and linguistic supervision.
Figure~\ref{fig:framework} summarizes the framework.

\subsection{Learning from Three-Level Text Supervision}
\label{sec:multilevel_learning}

\paragraph{Complementary supervision.}
The three caption levels link microscopy appearance to biological
semantics at complementary levels of specificity.
Category captions name the annotated structure--cell-line composition,
providing a common identity anchor for images in the same category.
Knowledge captions describe the biological properties of the annotated
entities, providing shared semantic context across compositions that
contain the same structure or cell line.
Detail captions describe image-specific fluorescence patterns,
morphology, signal intensity, and background, capturing visual variation
within a category.
Together, these text views allow images to be associated with both
shared biological meaning and instance-specific appearance.


\paragraph{Text-view sampling.}
Each time a microscopy image $x_i$ is selected for training, we sample
one caption to form the positive image--text pair $(x_i,t_i)$:
\begin{equation}
    t_i =
    \begin{cases}
        t_i^{\mathrm{cat}},  & \text{with probability } 0.50, \\
        t_i^{\mathrm{know}}, & \text{with probability } 0.25, \\
        t_i^{\mathrm{det}},  & \text{with probability } 0.25.
    \end{cases}
    \label{eq:caption_router}
\end{equation}
The higher sampling probability for category captions maintains frequent
supervision of target identity, while knowledge and detail captions provide
complementary context and visual descriptions. Captions are resampled on
each occurrence throughout training, allowing the same image to receive
different text views across repeated observations.

\paragraph{Contrastive learning.}
\label{sec:clip_preliminaries}
Let $\mathbf{v}_i$ and $\mathbf{u}_i$ be the $\ell_2$-normalized outputs of the image encoder $f_\theta(x_i)$ and text encoder $g_\phi(t_i)$. For a batch of $B$ pairs, the similarity logits are
\begin{equation}
    \ell_{ij}=\alpha\,\mathbf{v}_i^\top\mathbf{u}_j,
    \label{eq:clip_similarity}
\end{equation}
where $\alpha>0$ is a learned logit scale. We optimize the standard symmetric CLIP loss~\citep{radford2021learning}:
\begin{equation}
    \mathcal{L}_{\mathrm{CLIP}}=-\frac{1}{2B}\sum_{i=1}^{B}
    \left[
    \log\frac{\exp(\ell_{ii})}{\sum_{j=1}^{B}\exp(\ell_{ij})}
    +\log\frac{\exp(\ell_{ii})}{\sum_{j=1}^{B}\exp(\ell_{ji})}
    \right].
    \label{eq:clip_loss}
\end{equation}
All text views share the same encoder and objective, bringing the three forms of supervision into a common image--text space.

\subsection{Composition-Aware Sampling}
\label{sec:training_distribution}

Figure~\ref{fig:training_category_distribution} shows that the microscopy
training set follows a long-tailed distribution over structure--cell-line
compositions. To reduce the dominance of frequent compositions, we follow
the metadata-balancing strategy of MetaCLIP~\citep{xu2024demystifying}.
Each candidate occurrence associated with composition $e$ is retained with
probability

\begin{equation}
    p_{\mathrm{keep}}(e)
    =
    \min\left(1,\frac{\tau_a}{n(e)}\right),
    \label{eq:keep_probability}
\end{equation}

where $n(e)$ is the number of candidate records for composition
$e$, and $\tau_a=5{,}000$ controls the effective exposure of frequent
compositions. It retains all candidates from compositions below
the threshold while progressively reducing the contribution of those with
larger image counts.

We additionally apply standard rotation and reflection augmentation to
sparsely represented compositions before constructing the training stream.
The resulting augmented records are included in $n(e)$, and the
augmentation criterion and allocation procedure are provided in
Appendix~\ref{app:rare_class_augmentation}. Retention is applied before text-view
sampling, so the exposure of subcellular structure--cell-line compositions and the
selection of text levels are controlled independently.

\subsection{Auxiliary General-Domain Image--Text Data}
\label{sec:vlm_enrichment}

MicroPool-2M provides domain-specific microscopy supervision, but its image
and text distributions are narrower than those of
general-domain vision--language corpora. We therefore mix in VLM-0.7M, a
subset of 609,740 image--text records drawn from the VLM-150M dataset
released by HQ-CLIP~\citep{Wei_2025_ICCV}, to provide broader visual and
linguistic exposure during training.

Each auxiliary image is paired with its tag string or VLM-generated
description with equal probability. These records are mixed with
MicroPool-2M under the same contrastive objective, providing broader
visual and linguistic supervision.


\section{Experiments}
\subsection{Experimental Settings}
\label{sec:experimental_settings}

\begin{figure}[t]
\centering
\begin{minipage}[t]{0.60\textwidth}
    \vspace{0pt}
    \centering
    \begin{minipage}[c][5.2cm][c]{\linewidth}
        \centering
        \includegraphics[
            width=\linewidth,
            height=5.2cm,
            keepaspectratio
        ]{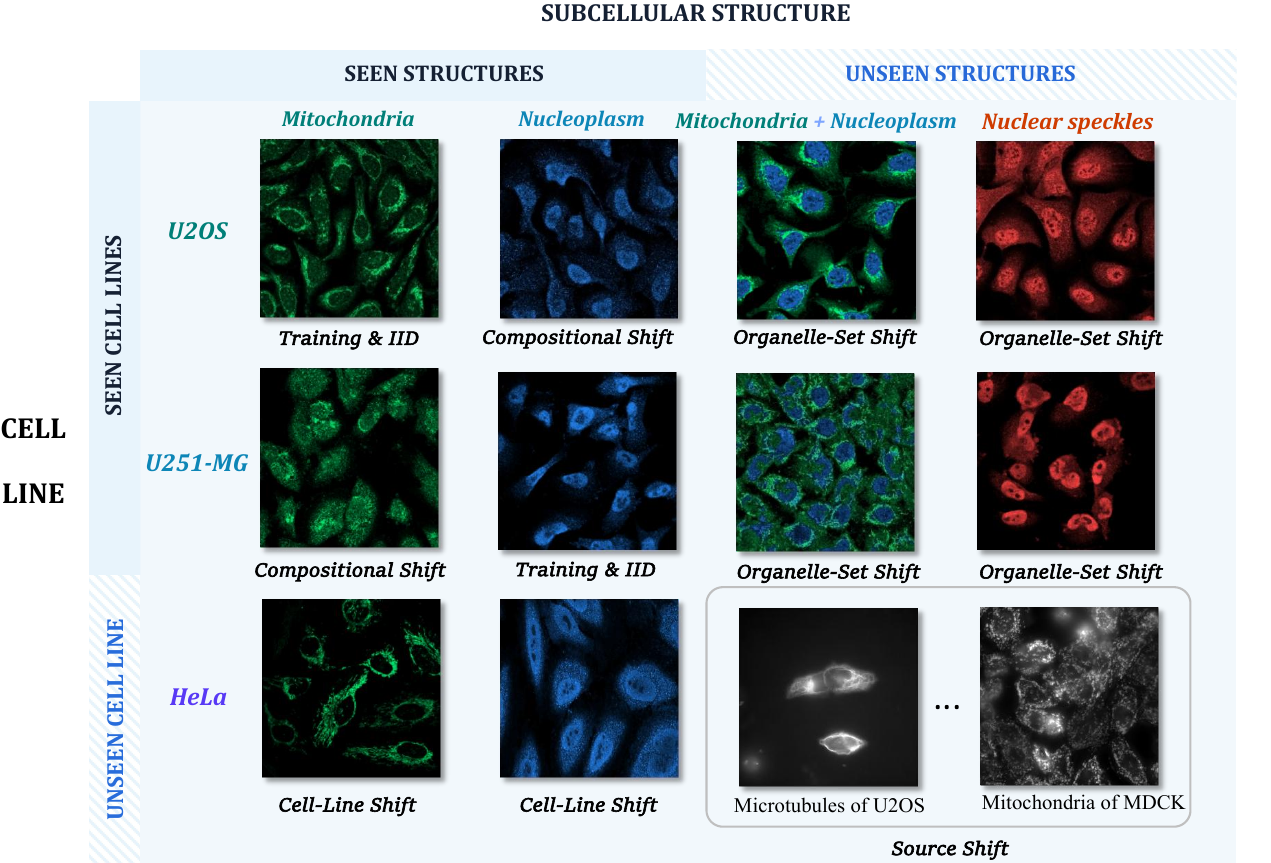}
    \end{minipage}
    \par\smallskip
    {\small (a) Evaluation examples.\par}
\end{minipage}\hfill
\begin{minipage}[t]{0.38\textwidth}
    \vspace{0pt}
    \centering
    \begin{minipage}[c][5.2cm][c]{\linewidth}
        \centering
        \footnotesize
        \setlength{\tabcolsep}{3pt}
        \renewcommand{\arraystretch}{1.5}
        \begin{tabular}{@{}lrr@{}}
            \toprule
            Subset & Classes & Images \\
            \midrule
            IID & 59 & 5,900 \\
            Compositional Shift & 64 & 6,400 \\
            Cell-Line Shift & 59 & 5,900 \\
            Organelle-Set Shift & 54 & 5,400 \\
            Source Shift (external) & 13 & 1,158 \\
            \midrule
            Training & 2,032 & 2,505,017 \\
            \bottomrule
        \end{tabular}
    \end{minipage}
    \par\smallskip
    {\small (b) Evaluation statistics.\par}
\end{minipage}
\caption{Classification evaluation subsets:
(a) representative images displayed using pseudocolor;
(b) category and image counts for the four internal splits,
the external Source Shift benchmark, and the training set.}
\label{fig:evaluation_examples}
\end{figure}

\paragraph{Evaluation data and protocols.}
We evaluate classification on four MicroPool-2M subsets defined relative to its microscopy training set and summarized in Table~\ref{tab:micropool}. \emph{IID} contains new images of observed structure--cell-line pairs. \emph{Compositional Shift} holds out pairs while retaining both constituent labels in training. \emph{Cell-Line Shift} holds out cell lines while retaining the structure labels, and \emph{Organelle-Set Shift} holds out complete structure labels while retaining the cell lines. The latter concerns structure labels, which can themselves be composite. Each internal evaluation category contains 100 images. We additionally evaluate \emph{Source Shift} using 1,158 images from 13 categories drawn from the external Light My Cells dataset (S-BIAD1047)~\citep{kauffmann2026lightmycells}. Figure~\ref{fig:evaluation_examples} shows representative images from these evaluation settings.

For retrieval, we use a held-out pool of 1,000 MicroPool image--text pairs, with each image paired with its detail-level caption. We additionally evaluate zero-shot transfer to projected 3D cell images from the Allen WTC-11 collection in Appendix~\ref{app:downstream_visualization}.


\paragraph{Evaluation metrics.}
For classification, image--text models rank all candidate categories within each evaluated subset using the category template in Section~\ref{sec:micropool}. We report Top-1 accuracy without adaptation to the evaluation subsets. For retrieval, we report Recall@1 for image-to-text (I2T) and text-to-image (T2I), ranking the full corresponding pool for each query.

\paragraph{Baselines.}
The fully supervised classifier is trained on the full microscopy
training set and evaluated on the 59-category IID subset. For Cell-DINO~\citep{moutakanni2025celldino}, we self-supervise the backbone on the same 2.5M microscopy training images and fit separate linear probes for structure and cell-line recognition; their predictions are combined for IID and Compositional Shift evaluation. CLIP~\citep{radford2021learning}, SigLIP~\citep{zhai2023sigmoid}, SigLIP~2~\citep{tschannen2025siglip}, PE-Core~\citep{bolya2025perception}, and DFN2B-CLIP~\citep{fang2024data} use released weights and native configurations, without microscopy adaptation or prompt ensembling. These comparisons measure direct transfer from general-purpose pretraining.

\paragraph{Implementation details.}


MicroCLIP initializes both encoders from DFN2B-CLIP ViT-B/16 and fine-tunes all parameters with $224\times224$ image inputs and a 256-token text context. The training stream combines 2,505,017 microscopy images, 609,740 natural-image records, and 175,516 transformed copies of rare-class microscopy images. We train for 12,000 updates using AdamW, a global batch size of 8,192, a peak learning rate of $4\times10^{-5}$, 2,500 warmup steps, cosine decay, and weight decay 0.2. Training uses bf16 mixed precision on eight NVIDIA RTX PRO 5000 GPUs with 72\,GB of memory each and takes approximately 55 hours. 

\subsection{Classification and Cross-Modal Retrieval}
\label{sec:main_results}

Table~\ref{tab:classification_retrieval} shows that MicroCLIP generalizes to unseen structure--cell-line pairings while maintaining high IID accuracy. It achieves 67.70\% on Compositional Shift, compared with 5.12\% for the strongest general-purpose VLM and 0.98\% for Cell-DINO under factor-wise linear probing. On IID images, it reaches 98.46\%, compared with 97.74\% for the fully supervised classifier.

Transfer also extends to held-out cell lines, structure labels, and an external source. MicroCLIP achieves 34.07\% on Cell-Line Shift and 46.13\% on Organelle-Set Shift, versus 7.63\% and 2.54\% for the strongest general-purpose baselines. Organelle-Set Shift concerns unseen complete labels, which may be composite, rather than necessarily unseen atomic structures. On Light My Cells~\citep{kauffmann2026lightmycells}, MicroCLIP reaches 36.53\%, versus 22.54\% for PE-Core, although accuracy under distribution shift remains well below IID performance.

The same representation supports retrieval with image-specific descriptions. On 1,000 held-out image--caption pairs with detail-level captions, MicroCLIP achieves 61.5\% I2T and 63.3\% T2I Recall@1, compared with at most 0.7\% and 0.3\% for the evaluated general-purpose VLMs.

\begin{table*}[t]
    \centering
    \caption{Comparison of classification and cross-modal retrieval performance. Classification results are Top-1 accuracy (\%), retrieval results are Recall@1 (\%). Best and second-best results are shown in bold and underlined, respectively.}
    \label{tab:classification_retrieval}
    \setlength{\tabcolsep}{3.5pt}
    \renewcommand{\arraystretch}{1.15}
    \resizebox{\textwidth}{!}{%
    \begin{tabular}{@{}lcc cccc cc@{}}
        \toprule
\multirow{2}{*}{Method} &
\multirow{2}{*}{Params.} &
\multicolumn{1}{c}{In-domain classification} &
\multicolumn{4}{c}{Out-of-domain classification} &
\multicolumn{2}{c}{Retrieval} \\
\cmidrule(lr){3-3}\cmidrule(lr){4-7}\cmidrule(l){8-9}
& & IID & Compositional Shift & Cell-Line Shift & Organelle-Set Shift & Source Shift &
I2T & T2I \\
        \midrule
        \multicolumn{9}{@{}l}{\textit{Closed-set methods}} \\
        \addlinespace[2pt]
        Fully Supervised Classifier
                            & --     & \underline{97.74} & --         & --           & --            & --            & --            & --            \\
        Cell-DINO          & 304M   & 36.06         & 0.98          & --           & --            & --            & --            & --            \\
        \midrule
        \multicolumn{9}{@{}l}{\textit{Open-set methods}} \\
        \addlinespace[2pt]
        CLIP (OpenAI)      & 150M   & 2.97          & \underline{5.12} & 1.80      & \underline{2.54} & 8.38        & \underline{0.7} & 0.2       \\
        SigLIP             & 203M   & 3.05          & 1.88          & 1.07         & 1.26          & 9.33           & 0.2           & 0.2           \\
        SigLIP 2           & 375M   & 3.08          & 2.28          & 3.59         & 1.11          & 10.79          & 0.2           & \underline{0.3} \\
        PE-Core            & 448M   & 3.34          & 3.78          & \underline{7.63} & 2.19       & \underline{22.54} & 0.4        & 0.0           \\
        DFN2B-CLIP         & 150M   & 1.44          & 1.41          & 4.32         & 0.81          & 16.49          & 0.5           & 0.1           \\
        \midrule
        \textbf{MicroCLIP (Ours)}
                            & 150M   & \textbf{98.46} & \textbf{67.70} & \textbf{34.07} & \textbf{46.13} & \textbf{36.53} & \textbf{61.5} & \textbf{63.3} \\
        \bottomrule
    \end{tabular}%
    }
\end{table*}

\subsection{Ablation Studies}
\label{sec:ablation_studies}

\paragraph{Ablation on Main Components.}

Table~\ref{tab:microclip_ablation} compares components using the same DFN2B-initialized ViT-B/16 backbone and contrastive objective. Category-level supervision supports compositional recognition but provides limited alignment with image-specific descriptions. Adding detail captions raises retrieval to 28.4\%/27.8\%, and adding knowledge captions further raises it to 37.1\%/37.4\%. Compositional Shift accuracy increases from 56.47\% to 66.36\%.

Composition balancing then improves Cell-Line Shift accuracy from 25.41\% to 31.53\%. With the auxiliary general-domain image--text pairs from VLM-0.7M, the full recipe reaches 61.5\%/63.3\% retrieval and the highest Compositional and Cell-Line Shift accuracy in this ablation. Gains vary across shifts: the balanced microscopy-only variant performs best on Organelle-Set Shift, while category-plus-detail supervision performs best on Source Shift.

\begin{table*}[t]
\centering
\caption{MicroCLIP component ablations. Cat./Det./Know.: category/detail/knowledge captions. Bal.: composition balancing. Aux.: general-domain image--text pairs from VLM-0.7M.}
\label{tab:microclip_ablation}
\resizebox{\textwidth}{!}{%
\begin{tabular}{lccccc c cccc cc}
\toprule
\multirow{2}{*}{Method} &
\multirow{2}{*}{Cat.} &
\multirow{2}{*}{Det.} &
\multirow{2}{*}{Know.} &
\multirow{2}{*}{Bal.} &
\multirow{2}{*}{Aux.} &
\multicolumn{1}{c}{In-domain classification} &
\multicolumn{4}{c}{Out-of-domain classification} &
\multicolumn{2}{c}{Retrieval} \\
\cmidrule(lr){7-7}\cmidrule(lr){8-11}\cmidrule(lr){12-13}
 & & & & & &
 IID & Compositional Shift & Cell-Line Shift & Organelle-Set Shift & Source Shift &
 I2T & T2I \\
\midrule
Category-level only
    & \checkmark &            &            &            &            
    & 97.37 & 56.47 & 23.25 & 40.74 & \underline{37.31} & 15.8  & 12.6  \\

+ Detail-level captions
    & \checkmark & \checkmark &            &            &            
    & \textbf{98.90} & 63.94 & 28.69 & 41.63 & \textbf{40.59} & 28.4 & 27.8 \\

+ Knowledge-level captions
    & \checkmark & \checkmark & \checkmark &            &            
    & \underline{98.86} & \underline{66.36} & 25.41 & \underline{48.31} & 36.36 & 37.1 & 37.4 \\

+ Data Balancing
    & \checkmark & \checkmark & \checkmark & \checkmark &            
    & 98.24 & 64.95 & \underline{31.53} & \textbf{49.91} & 36.44 & \underline{39.4} & \underline{41.5} \\
\rowcolor{gray!10}
+ VLM-0.7M
    & \checkmark & \checkmark & \checkmark & \checkmark & \checkmark
    & 98.46 & \textbf{67.70} & \textbf{34.07} & 46.13 & 36.53 & \textbf{61.5} & \textbf{63.3} \\
\bottomrule
\end{tabular}%
}
\end{table*}

\paragraph{Composition-balancing threshold.}
As shown in Table~\ref{tab:balancing_threshold}, lowering $\tau_a$ reduces the sampling dominance of frequent structure--cell-line pairs. Our default, $\tau_a=5{,}000$, improves Cell-Line Shift accuracy from 26.61\% without balancing to 34.07\% and gives the best Compositional Shift and bidirectional retrieval results among the tested thresholds, with little change in IID accuracy. The gains are not monotonic: Organelle-Set Shift favors $\tau_a=20{,}000$, while Source Shift favors $\tau_a=50{,}000$.

\begin{table*}[!htbp]
    \centering
    \caption{Effect of the composition-balancing threshold $\tau_a$ (Equation~\ref{eq:keep_probability}). Retained (\%) denotes the online sample acceptance fraction. Metrics and highlighting follow Table~\ref{tab:classification_retrieval}.}
    \label{tab:balancing_threshold}
    \setlength{\tabcolsep}{3.5pt}
    \renewcommand{\arraystretch}{1.15}
    \resizebox{\textwidth}{!}{%
    \begin{tabular}{@{}lcc cccc cc@{}}
        \toprule
        \multirow{2}{*}{Balancing threshold $\tau_a$} &
        \multirow{2}{*}{Retained (\%)} &
        \multicolumn{1}{c}{In-domain classification} &
        \multicolumn{4}{c}{Out-of-domain classification} &
        \multicolumn{2}{c}{Retrieval} \\
        \cmidrule(lr){3-3}\cmidrule(lr){4-7}\cmidrule(l){8-9}
        & & IID & Compositional Shift & Cell-Line Shift & Organelle-Set Shift & Source Shift &
        I2T & T2I \\
        \midrule

        off (no balancing) & 100.0
        & \underline{98.76}
        & \underline{66.64}
        & 26.61
        & 46.78
        & 30.05
        & 32.4
        & 35.7 \\

        $50{,}000$ & 49.8
        & \textbf{98.83}
        & 64.73
        & \underline{28.44}
        & \underline{47.35}
        & \textbf{39.29}
        & \underline{47.0}
        & \underline{47.6} \\

        $20{,}000$ & 30.7
        & 98.24
        & 65.52
        & 27.88
        & \textbf{47.80}
        & 34.54
        & 44.5
        & 45.5 \\

        \rowcolor{gray!10}
        $5{,}000$ & 10.5
        & 98.46
        & \textbf{67.70}
        & \textbf{34.07}
        & 46.13
        & \underline{36.53}
        & \textbf{61.5}
        & \textbf{63.3} \\

        \bottomrule
    \end{tabular}%
    }
\end{table*}

\paragraph{Training-data size.}
Table~\ref{tab:scale} shows that increasing microscopy data from 25\% to 100\% under an equal-epoch protocol improves Compositional Shift accuracy from 60.39\% to 66.97\% and strengthens retrieval, but reduces Cell-Line and Source Shift accuracy. Thus, more data benefits compositional recognition and retrieval without uniformly improving transfer. The final row reports the full recipe with VLM-0.7M~\citep{Wei_2025_ICCV}; it also changes balancing and rare-class augmentation, so it does not isolate the effect of additional data.

\begin{table*}[!htbp]
    \centering
\caption{Training-data scale and the full recipe. Samples (M) counts images before augmentation and balancing. Metrics and highlighting follow Table~\ref{tab:classification_retrieval}.}
    \label{tab:scale}
    \setlength{\tabcolsep}{3.5pt}
    \renewcommand{\arraystretch}{1.15}
    \resizebox{\textwidth}{!}{%
    \begin{tabular}{@{}lcc cccc cc@{}}
        \toprule
        \multirow{2}{*}{Training data} &
        \multirow{2}{*}{Samples (M)} &
        \multicolumn{1}{c}{In-domain classification} &
        \multicolumn{4}{c}{Out-of-domain classification} &
        \multicolumn{2}{c}{Retrieval} \\
        \cmidrule(lr){3-3}\cmidrule(lr){4-7}\cmidrule(l){8-9}
        & & IID & Compositional Shift & Cell-Line Shift & Organelle-Set Shift & Source Shift & I2T & T2I \\
        \midrule

        $25\%$ & 0.63
        & \textbf{98.76}
        & 60.39
        & \underline{33.05}
        & 44.56
        & \textbf{40.93}
        & 32.1
        & 32.0 \\

        $50\%$ & 1.25
        & 98.49
        & 65.05
        & 30.44
        & 45.19
        & \underline{39.81}
        & 32.4
        & 30.8 \\

        $100\%$ & 2.51
        & \underline{98.54}
        & \underline{66.97}
        & 30.05
        & \underline{45.54}
        & 37.31
        & \underline{60.1}
        & \underline{60.7} \\

        \rowcolor{gray!10}
        $100\%$ $+$ VLM-0.7M & 3.11
        & 98.46
        & \textbf{67.70}
        & \textbf{34.07}
        & \textbf{46.13}
        & 36.53
        & \textbf{61.5}
        & \textbf{63.3} \\

        \bottomrule
    \end{tabular}%
    }
\end{table*}

\subsection{Qualitative Visualization}
\label{sec:qualitative_visualization}

\begin{figure}[!htbp]
    \centering
    \includegraphics[width=0.95\textwidth]{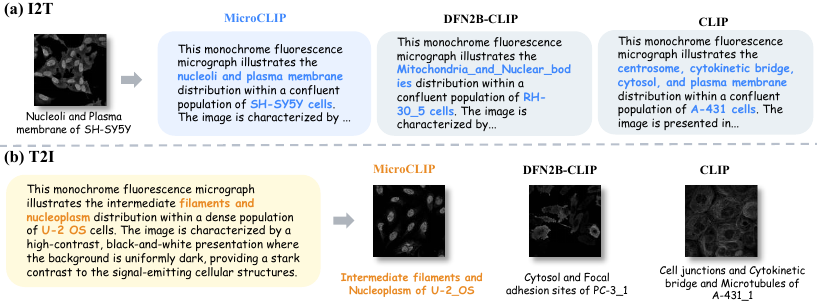}
    \caption{Qualitative comparison of (a) image-to-text and (b) text-to-image retrieval.}
    \label{fig:retrieval}
\end{figure}

\paragraph{Retrieval.}
To illustrate cross-modal retrieval, Figure~\ref{fig:retrieval} presents both retrieval directions. MicroCLIP retrieves descriptions and images matching the queried structure--cell-line identities, whereas DFN2B-CLIP and CLIP return mismatched annotations in these examples.

\begin{figure}[!htbp]
    \centering
    \includegraphics[width=0.95\textwidth]{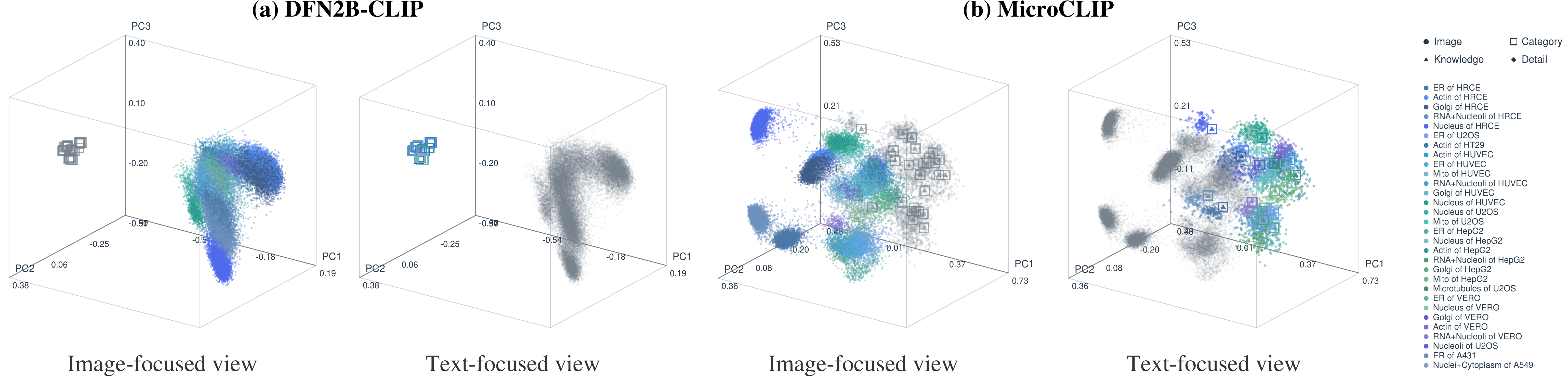}
    \caption{\textbf{Joint image--text embeddings visualized with 3D PCA.}
    (a) DFN2B-CLIP and (b) MicroCLIP. Each pair of views uses the same projection, highlighting images or text by category and showing the other modality in gray. Colors identify categories consistently across panels; marker shapes identify images and text levels. DFN2B-CLIP shows category text only, while MicroCLIP includes all three text levels.}
    \label{fig:3dpca}
\end{figure}

\paragraph{Joint image--text embedding space.}
Figure~\ref{fig:3dpca} shows that DFN2B-CLIP's category-text embeddings are separated from image embeddings, whereas MicroCLIP exhibits more distinct image clusters and greater overlap between modalities. This projection qualitatively illustrates the learned representation,
while classification and retrieval results quantify alignment.

\section{Limitations and conclusion}


MicroPool-2M provides three-level text supervision for MicroCLIP to connect fluorescence microscopy images with biological semantics. MicroCLIP outperforms the evaluated general-purpose vision--language models in zero-shot structure--cell-line recognition and bidirectional retrieval, supporting language-guided analysis beyond training categories. Our evaluation is limited to these tasks in fluorescence microscopy, while broader biological targets and research tasks remain outside the scope of this study. Extending language-guided microscopy analysis to these settings is a promising direction toward a more general-purpose microscopy research assistant.

\clearpage
\subsection*{AI use statement}
In this work, we used generative AI to generate synthetic text
annotations for MicroPool-2M. We also evaluated GPT-6 as a subject
of study in the qualitative examples and classification diagnostic
presented in Section~\ref{sec:observation} and
Appendix~\ref{app:grounding_diagnostic}.
We take full responsibility for the content of this work.

\subsection*{Ethics Statement}
MicroPool-2M is constructed from existing microscopy resources and
published data. We do not collect new human-subject data or personally
identifiable information. Data provenance, preprocessing, and source-specific
usage information are documented in the appendix and dataset documentation.

\subsection*{Reproducibility Statement}
To support reproducibility, we describe the construction and evaluation
splits of MicroPool-2M, as well as the MicroCLIP architecture, training
configurations, and evaluation protocols, in the main text and appendix.
We will publicly release MicroPool-2M, the MicroCLIP model weights, and
training code to facilitate reproduction of the reported results.

\bibliography{iclr2027_conference}
\bibliographystyle{iclr2027_conference}
\clearpage

\appendix
\makeatletter
\let\appendixOriginalAddcontentsline\addcontentsline
\renewcommand{\addcontentsline}[3]{%
  \appendixOriginalAddcontentsline{#1}{#2}{#3}%
  \def\appendixContentsType{#1}%
  \def\appendixTocType{toc}%
  \ifx\appendixContentsType\appendixTocType
    \addtocontents{apptoc}{%
      \protect\contentsline{#2}{#3}{\thepage}{\@currentHref}%
      \protected@file@percent}%
  \fi
}
\section*{Appendix Contents}
\begingroup
\setcounter{tocdepth}{2}
\@starttoc{apptoc}
\endgroup
\makeatother
\clearpage

\newcolumntype{C}[1]{>{\centering\arraybackslash}m{#1}}

\section{Additional Details of MicroPool-2M}
\label{app:micropool_details}


\subsection{Data Sources and Label Harmonization}
\label{app:data_sources}

MicroPool-2M draws its candidate microscopy images from public microscopy
datasets, including the Human Protein
Atlas (HPA)~\citep{thul2017subcellular}, the Image Data Resource
(IDR)~\citep{williams2017image}, RxRx~\citep{sypetkowski2023rxrx1}, the Broad
Bioimage Benchmark Collection (BBBC)~\citep{ljosa2012annotated}, and additional
public microscopy datasets. Table~\ref{tab:micropool_source_inventory} summarizes
the source datasets and their record counts. These counts
describe records in the source collection before MicroPool-specific
single-channel extraction, filtering, label harmonization, deduplication, and
split construction; they therefore should not be summed to recover the final
MicroPool-2M dataset size.

\begingroup
\footnotesize
\setlength{\tabcolsep}{4pt}
\setlength{\LTcapwidth}{\linewidth}
\begin{longtable}{
@{}C{15mm}| 
C{22mm}| 
C{23mm}| 
C{27mm}| 
C{18mm}| 
C{12mm}@{} 
}
\caption{Source datasets for MicroPool-2M. The table lists each dataset,
source publication, biological target, cell line or specimen, image dimensions, and record count. Values describe the collected
source data before MicroPool-specific extraction, filtering, harmonization, and
deduplication. For datasets summarized in a single row, the data number is the
sum over the listed structure--cell-line entries.} \label{tab:micropool_source_inventory}\\
\toprule
\hline
\textbf{Dataset} & \textbf{Publication} & \textbf{Subcellular Structure} & \textbf{Cell Line} & \textbf{Image Size} & \textbf{Data Number} \\
\hline
\endfirsthead

\multicolumn{6}{l}{\small\itshape Table \thetable\ (continued)}\\
\toprule
\hline
\textbf{Dataset} & \textbf{Publication} & \textbf{Subcellular Structure} & \textbf{Cell Line} & \textbf{Image Size} & \textbf{Data Number} \\
\hline
\endhead

\hline
\multicolumn{6}{r}{\small continue}\\
\endfoot
\endlastfoot

BBBC026 & Logan et al.~\cite{logan2016quantifying} & DNA & Human Hepatocyte and Murine Fibroblast cells & 1040$\times$1392 & 864 \\
\hline

\multirow{3}{16mm}{\centering BBBC001} & \multirow{3}{18mm}{\centering Moffat et al.~\cite{moffat2006lentiviral}} & DNA  & HT-29 & 512$\times$512 & 64512 \\
\cline{3-6}
& & Actin & HT-29 & 512$\times$512 & 64512 \\
\cline{3-6}
& & pH3 & HT-29 & 512$\times$512 & 64512 \\
\hline

\multirow{5}{16mm}{\centering BBBC020} & \multirow{5}{18mm}{\centering Wenzel et al.~\cite{wenzel2011measurement}} & CD11b & Bone Marrow-Derived Macrophages from C57BL/6 mice & 1388$\times$3$\times$1040 & 25 \\
\cline{3-6}
& & Nuclei & Bone Marrow-Derived Macrophages from C57BL/6 mice & 1388$\times$3$\times$1040 & 25 \\
\hline

\multirow{3}{16mm}{\centering BPAEC} & \multirow{3}{18mm}{\centering Zhang et al.~\cite{zhang2022correction}} & F-actin & BPAE & 1024$\times$1024 & 317 \\
\cline{3-6}
& & Mitochondria & BPAE & 1024$\times$1024 & 250 \\
\cline{3-6}
& & Nucleus & BPAE & 1024$\times$1024 & 111 \\
\hline

\multirow{4}{16mm}{\centering BioSR} & \multirow{4}{18mm}{\centering Qiao et al.~\cite{qiao2021evaluation}} & CCPs & COS-7 & 1004$\times$1004 & 54 \\
\cline{3-6}
& & ER & COS-7 & 1004$\times$1004 & 68 \\
\cline{3-6}
& & F-actin & COS-7 & 1004$\times$1004 & 51 \\
\cline{3-6}
& & Microtubules & COS-7 & 1004$\times$1004 & 55 \\
\hline

\multirow{5}{16mm}{\centering CPJUMP1} & \multirow{5}{18mm}{\centering Chandrasekaran et al.~\cite{chandrasekaran2024jump}} & AGP & U2OS & 1080$\times$1080 & 8896 \\
\cline{3-6}
& & DNA & U2OS & 1080$\times$1080 & 8896 \\
\cline{3-6}
& & ER & U2OS & 1080$\times$1080 & 8896 \\
\cline{3-6}
& & Mitochondria & U2OS & 1080$\times$1080 & 8898 \\
\cline{3-6}
& & RNA & U2OS & 1080$\times$1080 & 8897 \\
\hline

\multirow{2}{16mm}{\centering DeepBacs} & \multirow{2}{18mm}{\centering Spahn et al.~\cite{spahn2022deepbacs}} & H-NS-mScarlet-I & E.coli & 512$\times$512 & 28 \\

\cline{3-6}
& & MreB & E.coli & 512$\times$512 & 30 \\
\hline

\multirow{6}{16mm}{\centering Dynamic NuclearNet Segmentation} & \multirow{6}{18mm}{\centering Schwartz et al.~\cite{Schwartz803205}} & Nuclei & 3T3 & 512$\times$512 & 639 \\
\cline{3-6}
& & Nuclei & HEK293 & 512$\times$512 & 426 \\
\cline{3-6}
& & Nuclei & PC-3 & 512$\times$512 & 250 \\
\cline{3-6}
& & Nuclei & RAW264 & 512$\times$512 & 1755 \\
\cline{3-6}
& & Nuclei & RPE & 512$\times$512 & 200 \\
\cline{3-6}
& & Nuclei & HeLa & 512$\times$512 & 1680 \\
\cline{1-6}

\multirow{5}{16mm}{\centering Dynamic NuclearNet Tracking} & \multirow{5}{18mm}{\centering Schwartz et al.~\cite{Schwartz803205}} & Nuclei & 3T3 & 584$\times$600 & 639 \\
\cline{3-6}
& & Nuclei & HEK293 & 584$\times$600 & 426 \\
\cline{3-6}
& & Nuclei & PC-3 & 584$\times$600 & 250 \\
\cline{3-6}
& & Nuclei & RAW264 & 584$\times$600 & 1755 \\
\cline{3-6}
& & Nuclei & HeLa & 584$\times$600 & 1680 \\
\cline{3-6}
\hline

\multirow{3}{16mm}{\centering FMD} & \multirow{3}{18mm}{\centering Zhang et al.~\cite{zhang2019poisson}} & F-actin & BPAE & 512$\times$512 & 48 \\
\cline{3-6}
& & Mitochondria & BPAE & 512$\times$512 & 48 \\
\cline{3-6}
& & Nucleus & BPAE & 512$\times$512 & 48 \\
\cline{3-6}
\hline
\multirow{4}{16mm}{\centering Cell Tracking Challenge} & Bártová et al.~\cite{bartova2011recruitment} & GOWT1 & Mouse stem cells & 1024$\times$1024 & 184 \\
\cline{2-6}
& Ulman et al.~\cite{svoboda2016mitogen} & DNA & HL60 cells & 700$\times$1100 & 184 \\
\cline{2-6}
& Neumann et al.~\cite{neumann2010phenotypic} & H2B & HeLa cells & 690$\times$628 & 215 \\
\cline{2-6}
& Ivanov et al.~\cite{ivanov2012intestinal} & tia-1 & Human Hepatocarcinoma-derived Cells & 1024$\times$1024 & 60 \\
\hline

\multirow{10}{16mm}{\centering 2D HeLa} & \multirow{10}{18mm}{\centering Murphy et al.~\cite{murphy2000towards}} & Actin & HeLa & 382$\times$512 & 98 \\
\cline{3-6}
& & DNA & HeLa & 382$\times$512 & 862 \\
\cline{3-6}
& & Endosomes & HeLa & 382$\times$512 & 91 \\
\cline{3-6}
& & ER & HeLa & 382$\times$512 & 86 \\
\cline{3-6}
& & Golgi & HeLa, Giant & 382$\times$512 & 87 \\
\cline{3-6}
& & Golgi & HeLa, GPP130 & 382$\times$512 & 85 \\
\cline{3-6}
& & Lysosomes & HeLa & 382$\times$512 & 84 \\
\cline{3-6}
& & Microtubules & HeLa & 382$\times$512 & 91 \\
\cline{3-6}
& & Mitochondria & HeLa & 382$\times$512 & 73 \\
\cline{3-6}
& & Nucleolus & HeLa & 382$\times$512 & 80 \\
\hline

\multirow{4}{16mm}{\centering Caco-2} & \multirow{4}{18mm}{\centering Belharbi~\cite{belharbi2024sr}} & GFP-tubulin or E-cadherin & Caco-2 & 512$\times$512 & 9937 \\
\cline{3-6}
& & Survivin & Caco-2 & 512$\times$512 & 9937 \\
\cline{3-6}
& & mCherry-Histone H2B & Caco-2 & 512$\times$512 & 9937 \\
\hline

\multirow{2}{16mm}{\centering idr003} & \multirow{2}{18mm}{\centering Breker et al.~\cite{breker2013novel}} & GFP & Y6545 & 512$\times$672 & 83376 \\
\cline{3-6}
& & H2B-mCherry & Y6545 & 512$\times$672 & 83376 \\
\cline{1-6}

\multirow{3}{16mm}{\centering idr035} & \multirow{3}{18mm}{\centering Caie~\cite{caie2010high}} & Actin & MCF-7 & 1024$\times$1280 & 13200 \\
\cline{3-6}
& & Tubulin & MCF-7 & 1024$\times$1280 & 13200 \\
\cline{3-6}
& & DNA & MCF-7 & 1024$\times$1280 & 13200 \\
\cline{1-6}

\multirow{5}{16mm}{\centering idr036} & \multirow{5}{18mm}{\centering Gustafsdottir et al.~\cite{gustafsdottir2013multiplex}} & DNA & U2OS & 520$\times$696 & 17270 \\
\cline{3-6}
& & ER & U2OS & 520$\times$696 & 17272 \\
\cline{3-6}
& & Golgi and Actin & U2OS & 520$\times$696 & 13822 \\
\cline{3-6}
& & Nucleoli & U2OS & 520$\times$696 & 17272 \\
\cline{3-6}
& & Mitochondria & U2OS & 520$\times$696 & 13992 \\
\cline{1-6}

\multirow{3}{16mm}{\centering idr050} & \multirow{3}{18mm}{\centering Springer et al.~\cite{springer2019unbiased}} & Actin & HUVEC & 1488$\times$1488 & 3118 \\
\cline{3-6}
& & Microtubules & HUVEC & 1488$\times$1488 & 4310 \\
\cline{3-6}
& & Vimentin & HUVEC & 2048$\times$2048 & 1394 \\
\cline{1-6}

idr088 & Cox et al.~\cite{cox2020tales} & ACTB, CANX, CLTA, COX4I1, GM130, Nuclei/Cytoplasm, RAB5A, RELA, SQSTM1, TP53BP1, TUBA1B & A549, HepG2, WPMY-1 & 1050$\times$1250 & 102870 \\
\hline

\multirow{4}{16mm}{\centering idr093} & \multirow{4}{18mm}{\centering Muller~\cite{muller2021high}} & DNA & HeLa Kyoto & 2160$\times$2560 & 2973 \\
\cline{3-6}
& & Nascent RNA & HeLa Kyoto & 2160$\times$2560 & 2973 \\
\cline{3-6}
& & PCNA & HeLa Kyoto & 2160$\times$2560 & 2974 \\
\cline{3-6}
& & Succinimidyl Ester & HeLa Kyoto & 2160$\times$2560 & 2973 \\
\cline{1-6}

\multirow{7}{16mm}{\centering idr133} & \multirow{7}{18mm}{\centering Dahlin~\cite{dahlin2023reference}} & DNA & U2OS & 2160$\times$2160 & 648 \\
\cline{3-6}
& & ER & U2OS & 2160$\times$2160 & 759 \\
\cline{3-6}
& & F-actin and Plasma Membrane & U2OS & 2160$\times$2160 & 1250 \\
\cline{3-6}
& & RNA & U2OS & 2160$\times$2160 & 1092 \\
\cline{3-6}
& & Mitochondria & U2OS & 2160$\times$2160 & 940 \\
\hline

RxRx1 & Sypetkowski et al.~\cite{sypetkowski2023rxrx1} & Actin, ER, Golgi, Mitochondria, Nuclei, Nucleoli and Cytoplasmic RNA & HepG2, HUVEC, RPE & 512$\times$512 & 753060 \\
\hline
RxRx19a & Heiser~\cite{heiser2020identification} & Actin, ER, Golgi, Nuclei, Nucleoli and Cytoplasmic RNA & HRCE, Vero & 1024$\times$1024 & 1527600 \\
\hline

HPA & Ouyang et al.~\cite{ouyang2019analysis} & Nucleus, Microtubules, Nucleoplasm, ER, Cytosol, Mitochondria, Plasma Membrane, Nuclear Speckles, Golgi Apparatus, Intermediate Filaments, Cytokinetic Bridge, Nucleoli, Lipid Droplets, Nuclear Membrane, Actin Filaments, Microtubule Organizing Center, Cell Junctions, Centrosome, Nucleoli Fibrillar Center, Nuclear Bodies, Microtubules, Endoplasmic Reticulum, Rods $\&$ Rings, Mitotic Spindle, Aggresome, Focal Adhesion Sites, Endosomes, Lysosomes, Peroxisomes, Cytoplasmic bodies, Microtubule Ends & AF22, MCF7, U-251 MG, U2OS, CACO-2, A-431, HeLa, SH-SY5Y, RH-30, RT4, HEK 293, BJ, NIH 3T3, A549, PC-3, SK-MEL-30, HaCaT, SiHa, Hep G2, HUVEC TERT2, RPTEC TERT1, NB-4, REH, ASC TERT1, LHCN-M2, hTCEpi, HEL & 1728$\times$1728 & 279081 \\
\hline

idr037 & Vigilante et al.~\cite{vigilante2019identifying} & EdU, DNA, plasma & 62 induced pluripotent stem-cell lines and controls & 1360$\times$1024 & 46052 \\
\hline
\bottomrule
\end{longtable}
\endgroup

Source-specific names are normalized before defining a MicroPool category.
Aliases for the same cell line or structure are mapped to a canonical name, and
each retained image is assigned a complete structure label and a cell-line
identity. A complete structure label may contain one structure, one marker, or
a composition of several entries. Composite knowledge captions concatenate the available
constituent descriptions in canonical order.

\begin{table}[t]
\small
\centering
\caption{Classification evaluation data. Structure counts refer to complete structure labels, including composite labels. Each internal evaluation category contains 100 images. Source Shift is an external benchmark and is not balanced to 100 images per category.}
\label{tab:micropool}
\setlength{\tabcolsep}{5pt}
\begin{tabular}{lrrrrl}
\toprule
Subset & Classes & Struct. & Cells & Images & Shift \\
\midrule
IID & 59 & 21 & 12 & 5,900 & none \\
Compositional Shift & 64 & 16 & 16 & 6,400 & pair \\
Cell-Line Shift & 59 & 22 & 8 & 5,900 & cell line \\
Organelle-Set Shift & 54 & 29 & 6 & 5,400 & structure label \\
Source Shift (external) & 13 &  3 &  5 & 1,158 & source \\
\bottomrule
\end{tabular}
\end{table}
The final benchmark contains 2,505,017 training images and 23,600 internal
evaluation images. Each internal evaluation category contains 100 images.
Source Shift uses a separate external set of 1,158 images.
Table~\ref{tab:micropool} summarizes the category, structure, cell-line, and
image counts for each evaluation subset.
Splits are constructed from canonical structure--cell-line labels.
Cell-Line Shift holds out 8 cell lines, and Organelle-Set Shift holds out
29 complete structure labels, while retaining the other factor in training.
Compositional Shift selects 64 held-out pairs in ascending order of class size,
subject to each constituent structure and cell line remaining represented in
training. IID samples images from 59 categories that remain in training.
Rare-class augmentation is applied only to the training set.

\subsection{Training Category Distribution}
\label{app:category_distribution}

\begin{figure}[htbp]
\centering
\includegraphics[width=0.85\textwidth]{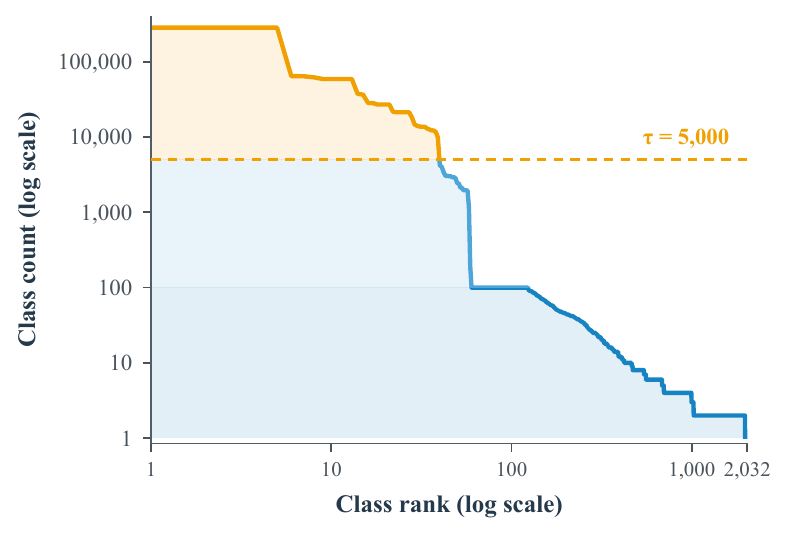}
\caption{Training category frequencies, ranked by image count. Dashed lines
mark the augmentation and retention thresholds.}
\label{fig:training_category_distribution}
\end{figure}

Figure~\ref{fig:training_category_distribution} shows the long-tailed frequency
distribution of the 2,032 training categories, each defined by a complete
structure--cell-line composition. Categories are ranked by decreasing image
count, with both axes shown on logarithmic scales. Under uniform image
sampling, frequent compositions receive more training exposure than sparsely
represented ones. This imbalance motivates the composition-aware sampling
in Section~\ref{sec:training_distribution}, which reduces the retention
probability of compositions with candidate counts above $\tau_a=5{,}000$.
The reference line marks this retention threshold; details of the training
stream and supplementary rare-class augmentation are provided in
Appendix~\ref{app:rare_class_augmentation}.

\section{Caption Construction}
\label{app:caption_construction}

This section documents how the three text views are constructed and provides
the prompts used to generate the knowledge-level and detail-level captions. It then
shows representative image--caption examples to illustrate the information
contributed by each view.

\subsection{Generation Procedure and Prompts}
\label{app:caption_prompts}
\begin{figure}[t]
\centering
\includegraphics[width=0.95\textwidth]{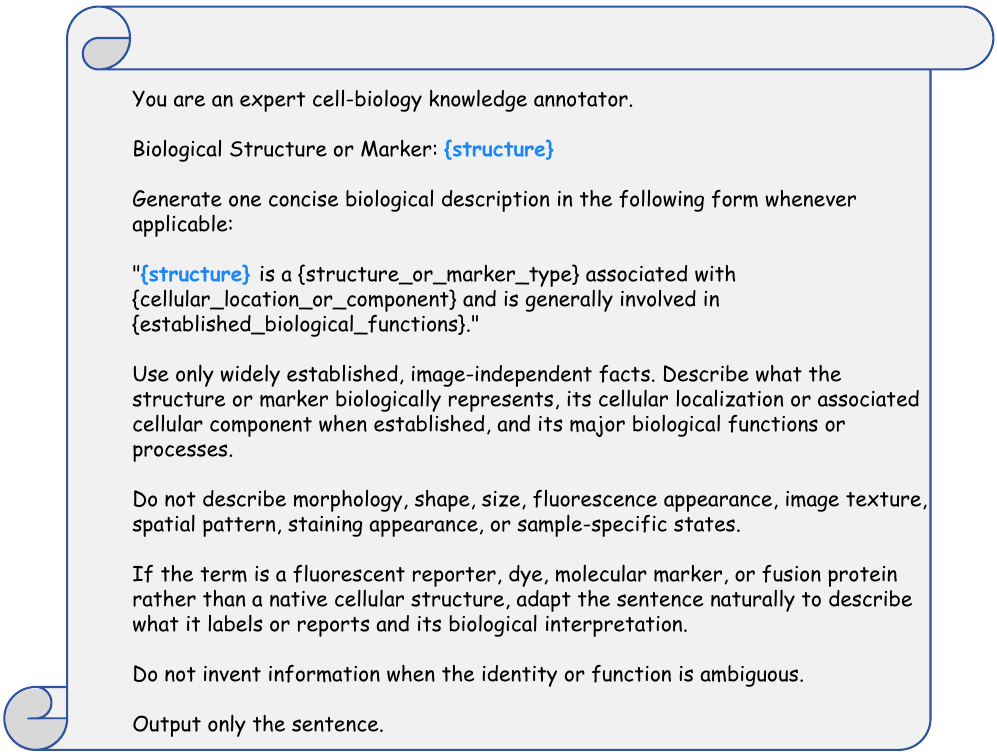}
\caption{Prompt for generating image-independent biological knowledge about subcellular structures and markers with GPT.}
\label{fig:prompt_knowledge_structure}
\end{figure}

\begin{figure}[t]
\centering
\includegraphics[width=0.95\textwidth]{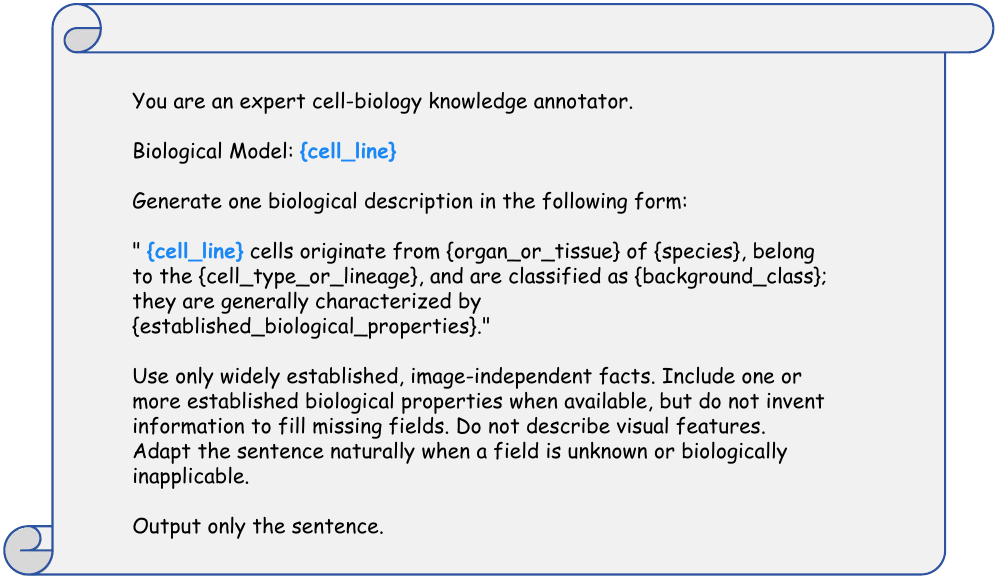}
\caption{Prompt for generating image-independent biological knowledge about cell lines with GPT.}
\label{fig:prompt_knowledge_cell}
\end{figure}

\begin{figure}[t]
\centering
\includegraphics[width=0.95\textwidth]{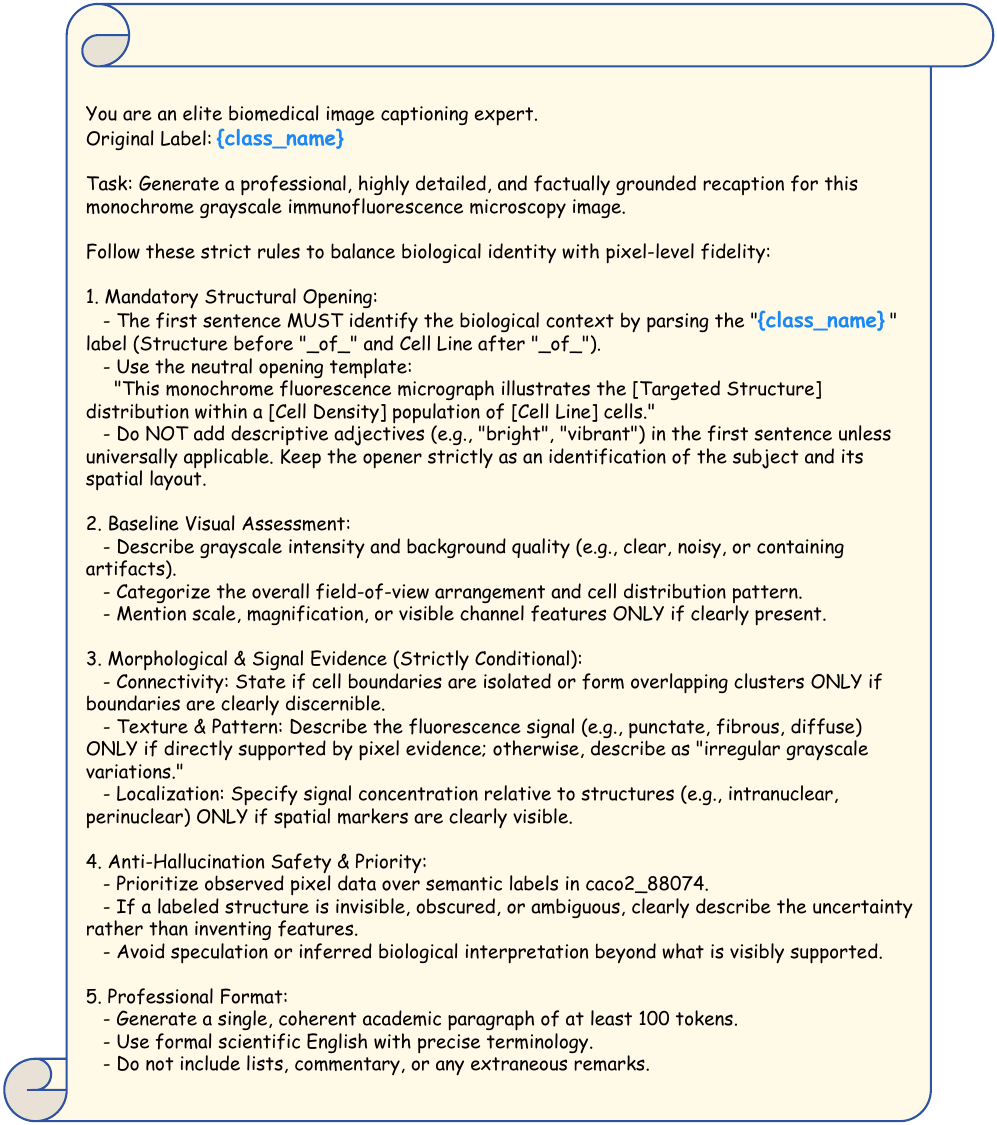}
\caption{Prompt for generating detail-level captions with Qwen from a microscopy image and its category annotation.}
\label{fig:prompt_detail}
\end{figure}

The category caption uses the deterministic template in Section~\ref{sec:micropool}. Knowledge captions are constructed from separate structure and cell-line descriptions generated with GPT-5.6 Sol. Detail captions are generated with Qwen3.5-397B-A17B conditioned on the image and its normalized category name. Figures~\ref{fig:prompt_knowledge_structure} and~\ref{fig:prompt_knowledge_cell} provide the knowledge prompts, and Figure~\ref{fig:prompt_detail} provides the detail prompt. The three resulting text views are illustrated in Figure~\ref{fig:three-level}.

\subsection{Examples of Three-Level Captions}
\label{app:caption_examples}

\begin{figure}[htbp]
\centering
\includegraphics[width=0.95\textwidth]{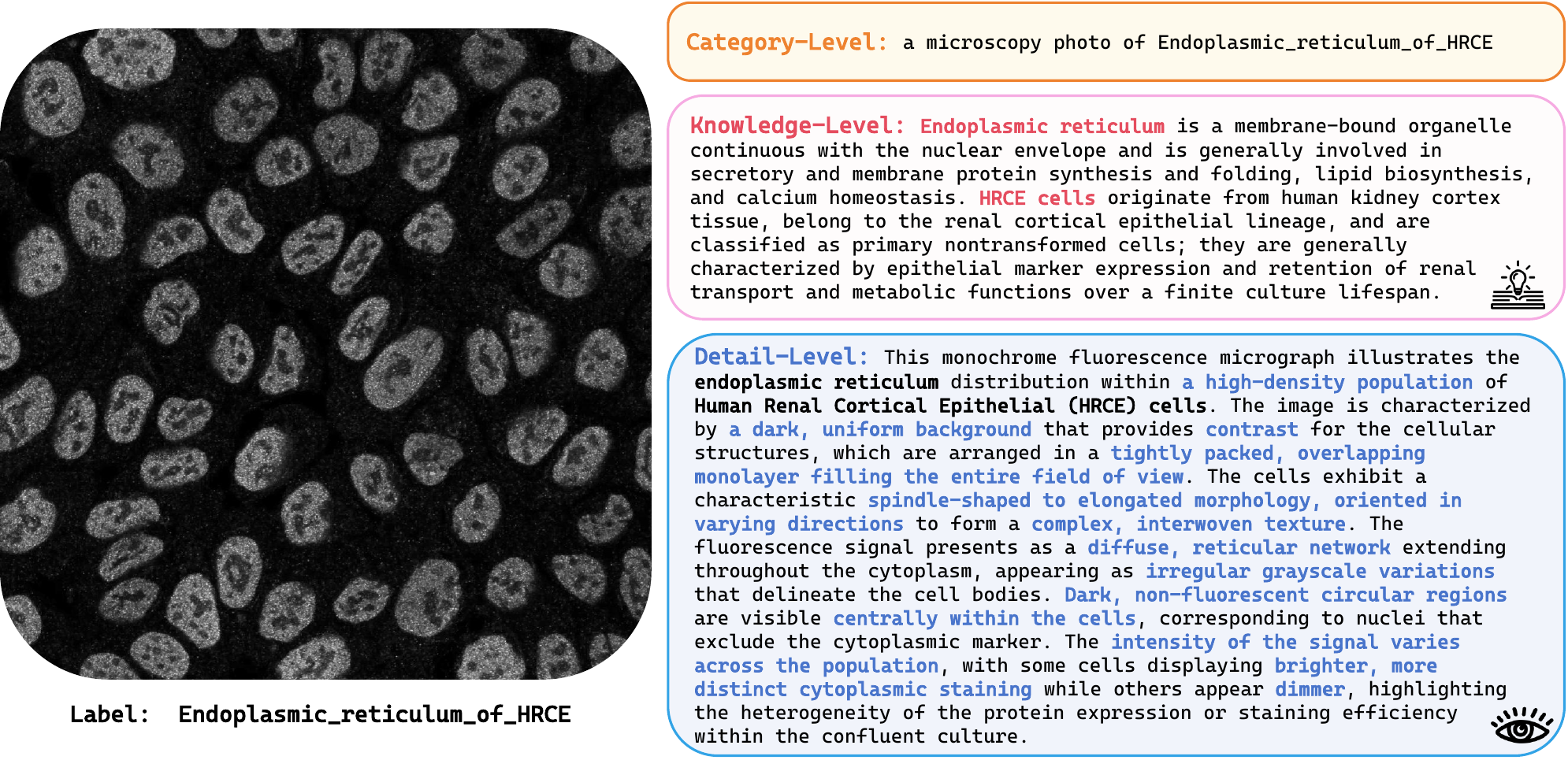}
\caption{Three-level supervision for one fluorescence microscopy image. The
category caption identifies the annotated structure and cell line; the
knowledge caption describes image-independent biological properties of these
entities; the detail caption describes the individual image.}
\label{fig:three-level}
\end{figure}

Figure~\ref{fig:three-level} illustrates how the three caption levels complement
one another for the same image. Category and knowledge captions provide a
shared semantic reference for images with the same annotation, whereas detail
captions describe image-specific appearance. Together, these views connect
category identity and biological context with the visual variation within a
category.

\section{Additional Training Details}
\label{app:training_details}

\subsection{Text Length and Tokenization}
\label{app:text_lengths}

\begin{table}[t]
\centering
\small
\caption{Token-length statistics for the three caption levels.}
\label{tab:caption_lengths}
\setlength{\tabcolsep}{8pt}
\renewcommand{\arraystretch}{1.15}
\begin{tabular}{@{}lrrrr@{}}
\toprule
Caption level & Median & P95 & Truncated at 77 (\%) & Truncated at 256 (\%) \\
\midrule
Category  & 15  & 19  & 0.00   & 0.00 \\
Detail    & 225 & 270 & 99.83  & 12.97 \\
Knowledge & 109 & 147 & 100.00 & $<0.01$ \\
\bottomrule
\end{tabular}
\end{table}

Table~\ref{tab:caption_lengths} summarizes the token lengths of
the three caption levels. Knowledge-level lengths are measured after
concatenating the descriptions of all constituent structures
and the cell line. We use the checkpoint's byte-level BPE
tokenizer and count tokens before truncation, excluding the
two special tokens.

Because the tokenizer adds start-of-text and end-of-text tokens,
a context of $N$ tokens accommodates up to $N-2$ caption tokens.
At the standard 77-token context, all knowledge-level captions
and 99.83\% of detail-level captions would be truncated.
We therefore use a 256-token text context, which accommodates 87.03\% of detail-level captions without
truncation. For knowledge-level captions, only 36 samples exceed
this limit and are truncated. Longer captions are truncated
during tokenization.

\subsection{Rare-Class Augmentation}
\label{app:rare_class_augmentation}

To expose the model more often to rare categories during training,
we augment categories with fewer than 100 images using rotations
and reflections. Each category corresponds to a canonical
structure--cell-line composition. Let $c_e$ denote the number of
original training images in category $e$. The number of additional
copies is
\begin{equation}
    a_e =
    \begin{cases}
        100-c_e, & c_e < 100,\\
        0, & c_e \geq 100.
    \end{cases}
    \label{eq:rare_class_augmentation}
\end{equation}
This brings each rare category to 100 training records and leaves
the other categories unchanged. In total, we add 175,516 records
to 1,909 rare categories containing 15,384 original images.

Within each rare category, we distribute the additional copies
evenly across the original images, with copy counts differing by
at most one. Any remainder is assigned in source record order.
For each image, we cycle through seven transformations: rotations
by $90^\circ$, $180^\circ$, and $270^\circ$, and reflections across
the horizontal, vertical, and two diagonal axes. The cycle repeats
when more than seven copies are needed. Each source image therefore
contributes at most seven distinct transformed views, and the
100 records per category may include repeated views.

Each augmented record retains the captions and metadata of the
source image. We combine the original and augmented records before
training and include both in $n(e)$ in
Equation~\ref{eq:keep_probability}. Both follow the same retention
and text-view sampling procedures.

\section{Biological Grounding Diagnostic}
\label{app:grounding_diagnostic}

\paragraph{Sampling and task.}
We complement the qualitative observation in Section~\ref{sec:observation}
with a 60-image classification diagnostic on selected subsets of IID and
Compositional Shift. Each subset contains 10 structure--cell-line
categories with three images per category. Classification uses a fixed
set of 10 candidate categories for each split, yielding a chance accuracy
of 10\%. The split definitions refer to MicroPool's training set.

\paragraph{Inputs and prediction.}
GPT-6 receives each image together with the candidate category labels
and is instructed to select the most likely structure--cell-line
combination, even when uncertain, and return only its candidate ID.
MicroCLIP and DFN2B-CLIP are evaluated in the same 10-way classification
setting. A prediction is correct only when both the structure and cell
line match the reference annotation.

\paragraph{Results and scope.}
Table~\ref{tab:grounding_diagnostic} reports joint structure--cell-line
accuracy. GPT-6 correctly identifies 22 of the 60 images (36.67\%),
compared with 10 (16.67\%) for DFN2B-CLIP and 57 (95.00\%) for MicroCLIP.
These results support the qualitative observation in
Figure~\ref{fig:observation}. This small diagnostic complements the full
benchmark evaluations but does not isolate the causes of the grounding gap.

\begin{table}[htbp]
\centering
\small
\caption{Sampled biological grounding diagnostic. Each split contains
10 categories with three images per category. Entries give correct
predictions / images and joint Top-1 accuracy. MicroCLIP is evaluated
after 15 training epochs.}
\label{tab:grounding_diagnostic}
\setlength{\tabcolsep}{7pt}
\begin{tabular}{lccc}
\toprule
Model & IID & Compositional Shift & Overall \\
\midrule
GPT-6 & 12/30 (40.00\%) & 10/30 (33.33\%) & 22/60 (36.67\%) \\
DFN2B-CLIP & 6/30 (20.00\%) & 4/30 (13.33\%) & 10/60 (16.67\%) \\
MicroCLIP & 30/30 (100.00\%) & 27/30 (90.00\%) & 57/60 (95.00\%) \\
\bottomrule
\end{tabular}
\end{table}

\section{Additional Quantitative Results}
\label{app:additional_quantitative_results}

\subsection{Model Scaling}
\label{app:model_scaling}
Among the PE-Core variants in Table~\ref{tab:backbone_scale}, PE-Core-G achieves the highest accuracy on all classification evaluations, with the largest gains over PE-Core-B on Compositional Shift and Organelle-Set Shift. Retrieval instead peaks at PE-Core-L and declines with PE-Core-G, indicating that scaling benefits classification more than retrieval in this comparison.

\begin{table*}[t]
    \centering
    \caption{
    MicroCLIP performance across backbone architectures and model sizes.
    Classification is zero-shot Top-1 accuracy (\%), and retrieval is Recall@1 (\%) 
    for image-to-text (I2T) and text-to-image (T2I) retrieval.
    Best and second-best results are shown in \textbf{bold} and \underline{underline}, respectively.
    }
    \label{tab:backbone_scale}
    \setlength{\tabcolsep}{3.5pt}
    \renewcommand{\arraystretch}{1.15}
    \resizebox{\textwidth}{!}{%
    \begin{tabular}{@{}lcc cccc cc@{}}
        \toprule
        \multirow{2}{*}{Backbone} &
        \multirow{2}{*}{Params.} &
        \multicolumn{1}{c}{In-domain classification} &
        \multicolumn{4}{c}{Out-of-domain classification} &
        \multicolumn{2}{c}{Retrieval} \\
        \cmidrule(lr){3-3}
        \cmidrule(lr){4-7}
        \cmidrule(l){8-9}
        & & IID & Compositional Shift & Cell-Line Shift &
        Organelle-Set Shift & Source Shift & I2T & T2I \\
        \midrule

        PE-Core-B/16
        & 448M
        & 98.80
        & 40.94
        & 24.27
        & 25.24
        & 21.68
        & 59.9
        & 61.5 \\

        PE-Core-L/14-336
        & 671M
        & \underline{99.07}
        & 49.44
        & 17.17
        & 32.22
        & 21.93
        & \underline{60.5}
        & \underline{61.8} \\

        PE-Core-G/14-448
        & 2419M
        & \textbf{99.20}
        & \textbf{74.34}
        & \underline{29.86}
        & \textbf{54.70}
        & \underline{35.06}
        & 54.7
        & 54.5 \\

        \addlinespace[2pt]

        \bottomrule
    \end{tabular}%
    }
\end{table*}

\subsection{Caption Sampling Ratios}
\label{app:caption_sampling_ratios}

Table~\ref{tab:text_ratio_ablation} examines how the choice and sampling
proportions of category-level (Cat.), detail-level (Det.), and knowledge-level
(Know.) captions relate to classification and retrieval performance. All runs
use a 256-token text context. Classification is measured by Top-1
accuracy, with Source Shift evaluated on Light-My-Cells. Retrieval is measured
by bidirectional Recall@1 on 1,000 held-out MicroPool image--text pairs with
detail-level captions.
The detail-only, knowledge-only, and category-plus-knowledge runs share the same
training setup apart from their caption sampling ratios. The category-only and
category-plus-detail runs use an earlier pipeline, while the three-level mix
uses the full MicroCLIP training recipe.

The single-level results reveal different performance profiles. Detail-only
supervision reaches 48.10\% I2T and 48.50\% T2I Recall@1, but achieves only
56.53\% IID and 26.53\% Compositional Shift accuracy. Knowledge-only supervision
also yields low classification accuracy and performs poorly on retrieval with
detail-level captions. Mixing category and knowledge captions raises IID
accuracy from 39.95\% to 97.47\% and Compositional Shift accuracy from 12.08\%
to 63.17\% relative to knowledge-only training. This comparison supports
retaining direct category supervision for recognition. Classification uses
category-name prototypes, so the single-level models trained without category
captions also face a mismatch between training descriptions and evaluation
text.

The full MicroCLIP configuration samples category, detail, and knowledge
captions with probabilities of 0.50, 0.25, and 0.25, respectively. It achieves
the highest Compositional Shift accuracy, Cell-Line Shift accuracy, and retrieval Recall@1 among the
reported configurations. It does not lead on every classification subset.
Category-plus-detail supervision gives the highest IID and Source Shift
accuracy, while category-plus-knowledge supervision performs best on
Organelle-Set Shift. These results show that the full configuration combines
strong recognition and retrieval performance, but the differences in training
pipelines mean that its gains cannot be attributed to the sampling ratio alone.

\begin{table*}[t]
\centering
\caption{Effect of caption sampling ratios. We compare single-level and mixed
supervision on classification and cross-modal retrieval.}
\label{tab:text_ratio_ablation}
\setlength{\tabcolsep}{3.5pt}
\renewcommand{\arraystretch}{1.15}
\resizebox{\textwidth}{!}{%
\begin{tabular}{@{}lccc c cccc cc@{}}
\toprule
\multirow{2}{*}{Method} &
\multicolumn{3}{c}{Sampling ratio} &
\multicolumn{1}{c}{In-domain classification} &
\multicolumn{4}{c}{Out-of-domain classification} &
\multicolumn{2}{c}{Retrieval} \\
\cmidrule(lr){2-4}\cmidrule(lr){5-5}\cmidrule(lr){6-9}\cmidrule(l){10-11}
 & Cat. & Det. & Know. & IID & Compositional Shift & Cell-Line Shift & Organelle-Set Shift & Source Shift & I2T & T2I \\
\midrule
Category-level only            & 1.00 & 0.00 & 0.00 & 97.37  & 56.47  & 23.25  & 40.74  & \underline{37.31} & 15.80   & 12.60   \\
Detail-level only              & 0.00 & 1.00 & 0.00 & 56.53  & 26.53  & 10.56  & 26.72  & 10.28  & \underline{48.10} & \underline{48.50} \\
Knowledge-level only           & 0.00 & 0.00 & 1.00 & 39.95  & 12.08  & 14.88  & 12.00  & 30.31  & 1.30  & 0.10  \\
Category $+$ Detail            & 0.75 & 0.25 & 0.00 & \textbf{98.90} & \underline{63.94} & 28.69  & 41.63  & \textbf{40.59} & 28.40  & 27.80  \\
Category $+$ Knowledge         & 0.75 & 0.00 & 0.25 & 97.47 & 63.17 & \underline{29.31} & \textbf{46.78} & 27.72 & 14.90 & 9.90 \\
\rowcolor{gray!12}
Category $+$ Detail $+$ Knowledge & 0.50 & 0.25 & 0.25 & \underline{98.46} & \textbf{67.70} & \textbf{34.07} & \underline{46.13} & 36.53 & \textbf{61.50} & \textbf{63.30} \\
\bottomrule
\end{tabular}%
}
\end{table*}

\section{Downstream Transfer Evaluation}
\label{app:downstream_visualization}

\paragraph{3D cell classification.}
We evaluate zero-shot subcellular-structure classification on 160 Allen WTC-11 hiPSC volumes~\citep{viana2023integrated}, with 32 volumes per class across actin filaments, endoplasmic reticulum, Golgi apparatus, microtubules, and mitochondria. Each volume comes from a distinct field of view. After per-volume percentile normalization ($p_{0.1}$--$p_{99.9}$), we obtain a 2D maximum-intensity projection (MIP):
\begin{equation}
I_{\mathrm{MIP}}(x,y)=\max_z I(z,x,y),
\label{eq:3dcell_mip}
\end{equation}
where $I$ is the normalized volume and $z$ indexes axial depth. All models classify the same projections by image--text similarity to the five class names, without task-specific training.

\begin{figure}[t]
\centering
\includegraphics[width=\textwidth]{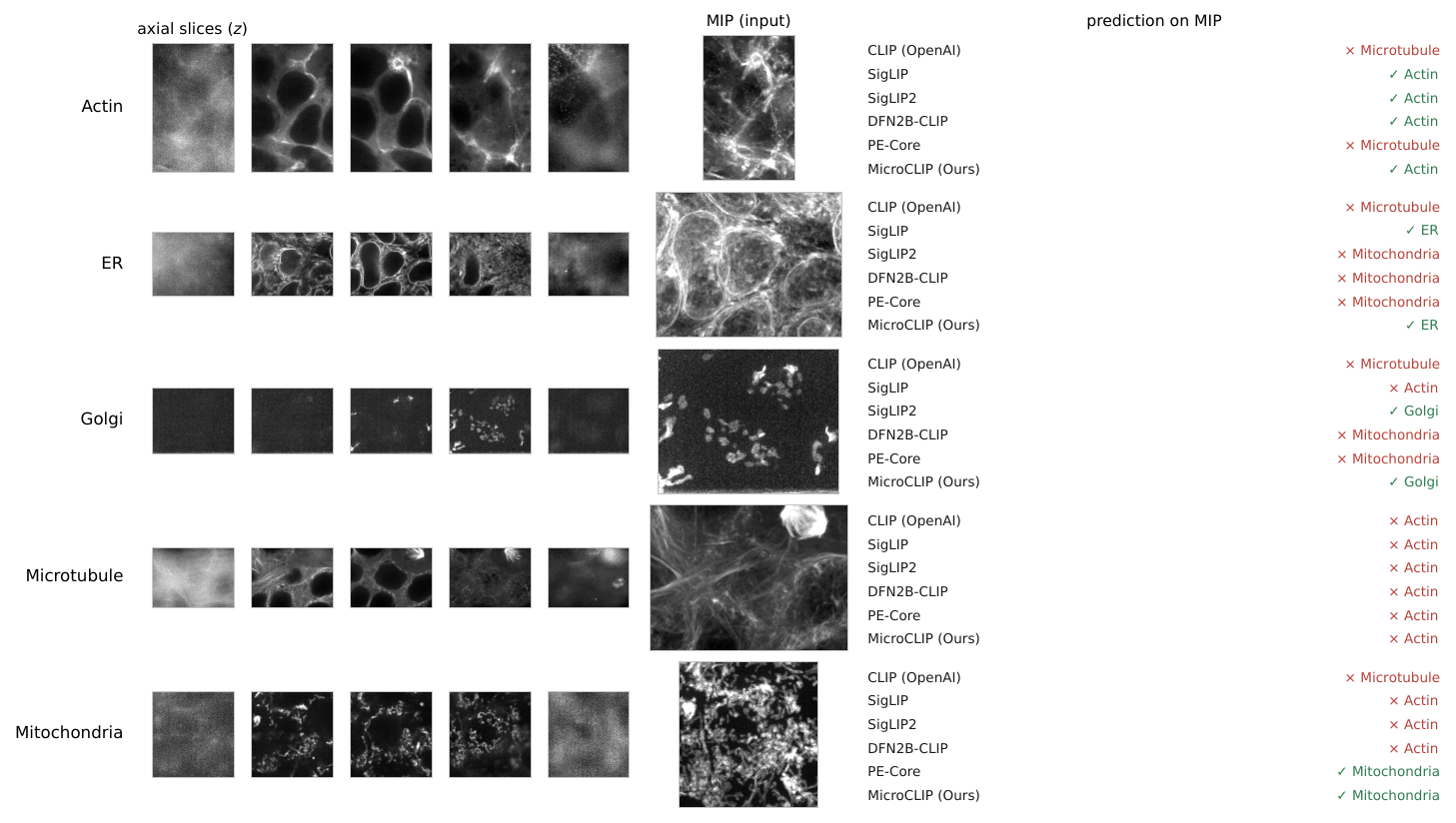}
\caption{3D cell classification examples. Each row shows five axial slices, the maximum-intensity projection used as model input, and predictions from six models. Green and red indicate correct and incorrect predictions, respectively. All models misclassify the displayed microtubule example as actin.}
\label{fig:3dcell_qualitative}
\end{figure}

Table~\ref{tab:3dcell} reports results under this MIP protocol. MicroCLIP achieves 88.12\% accuracy, exceeding the strongest baseline, PE-Core, by 46.25 percentage points. This demonstrates transfer to projected 3D cell images using the pretrained 2D encoder. Figure~\ref{fig:3dcell_qualitative} shows predictions for each class, including a microtubule example misclassified by all models.

\begin{table}[t]
    \centering
    \caption{Zero-shot classification of five subcellular structures on 160 Allen WTC-11 volumes using maximum-intensity projections. All models receive the same 2D projections; chance accuracy is 20\%.}
    \label{tab:3dcell}
    \small
    \setlength{\tabcolsep}{5pt}
    \renewcommand{\arraystretch}{1.08}
    \begin{tabular}{@{}lc@{}}
        \toprule
        Model & Top-1 accuracy (\%) \\
        \midrule
        CLIP (OpenAI) ViT-B/16    & 12.50 \\
        SigLIP ViT-B/16           & 22.50 \\
        SigLIP2 ViT-B/16          & 18.75 \\
        DFN2B-CLIP ViT-B/16       & 17.50 \\
        PE-Core B/16               & \underline{41.88} \\
        MicroCLIP (Ours)          & \textbf{88.12} \\
        \bottomrule
    \end{tabular}
\end{table}

\section{Visualizations and Downstream Results}
\label{app:qualitative_results}

\subsection{Classification Examples}
\label{app:classification_visualization}

\begin{figure}[htbp]
\centering
\includegraphics[width=0.95\textwidth]{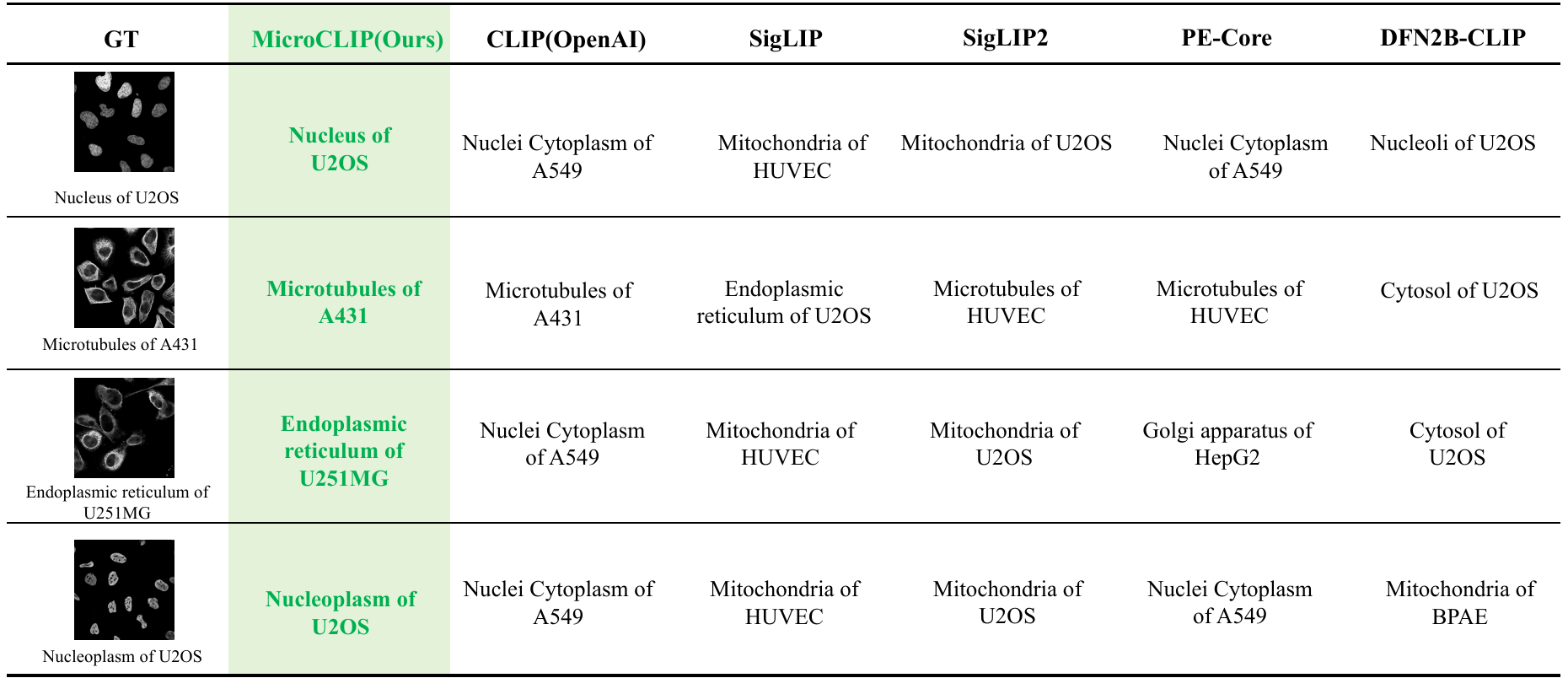}
\caption{IID classification examples comparing MicroCLIP and baseline predictions with ground-truth labels on new images of structure--cell-line pairs observed during training.}
\label{fig:IID}
\end{figure}

\begin{figure}[htbp]
\centering
\includegraphics[width=0.95\textwidth]{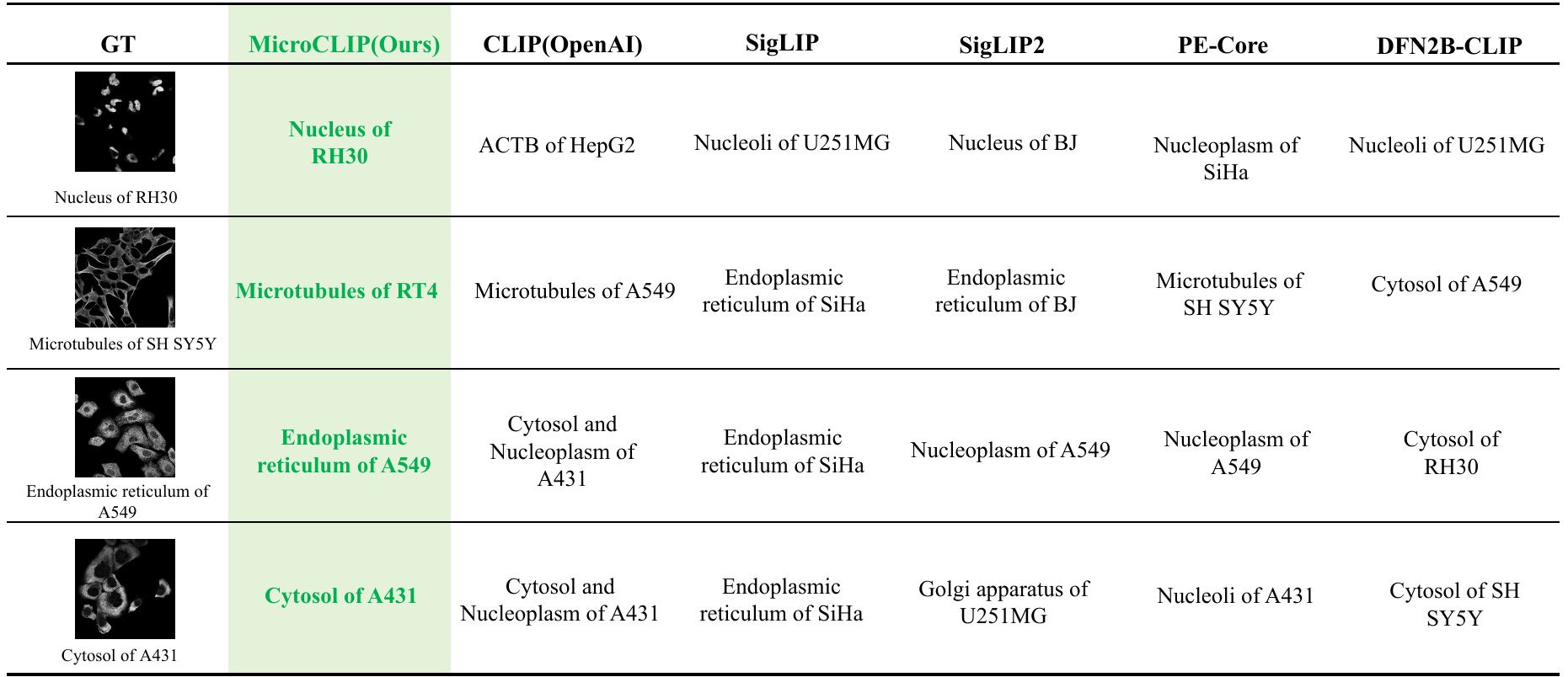}
\caption{Compositional Shift examples comparing model predictions on held-out structure--cell-line pairs whose individual constituents are represented in training.}
\label{fig:composition}
\end{figure}

\begin{figure}[htbp]
\centering
\includegraphics[width=0.95\textwidth]{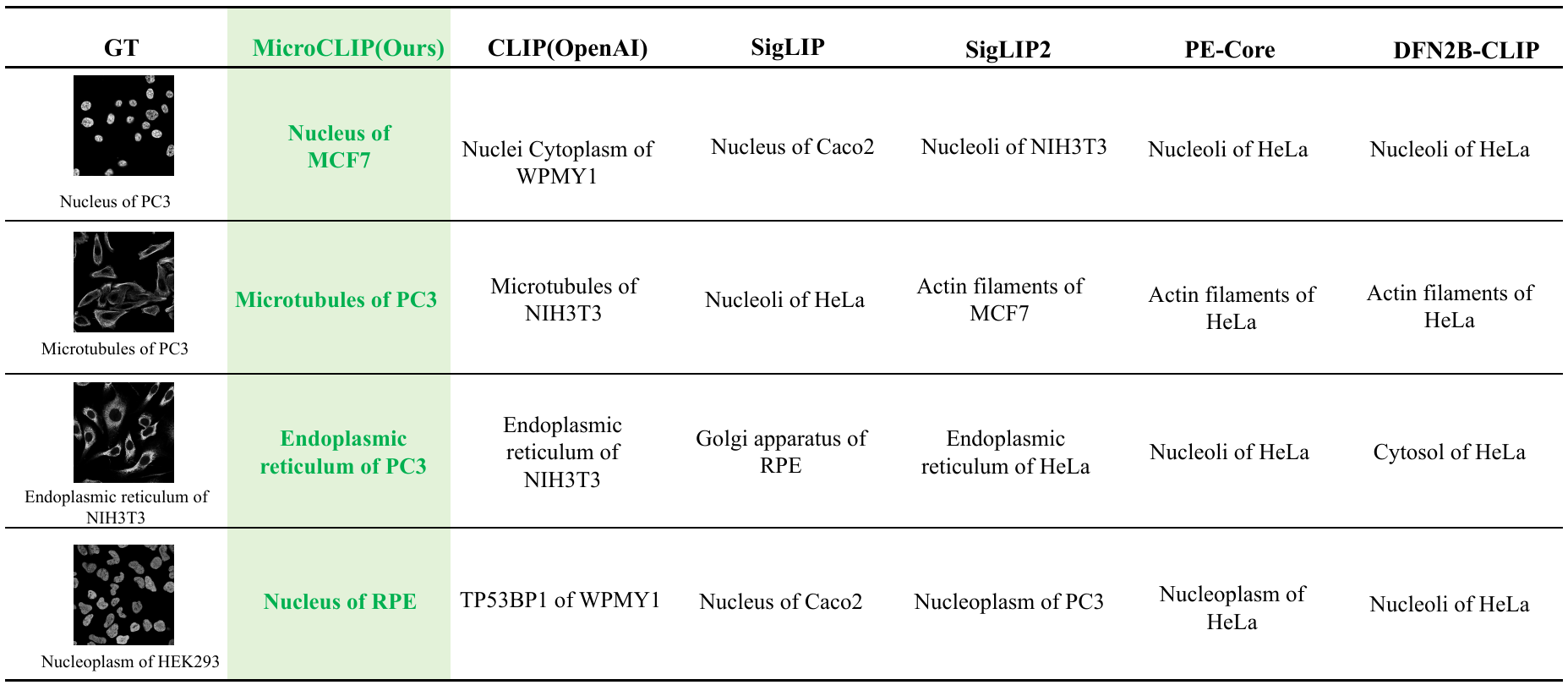}
\caption{Cell-Line Shift examples comparing model predictions on cell lines absent from training, with structure labels represented in training.}
\label{fig:cell_line}
\end{figure}

\begin{figure}[htbp]
\centering
\includegraphics[width=0.95\textwidth]{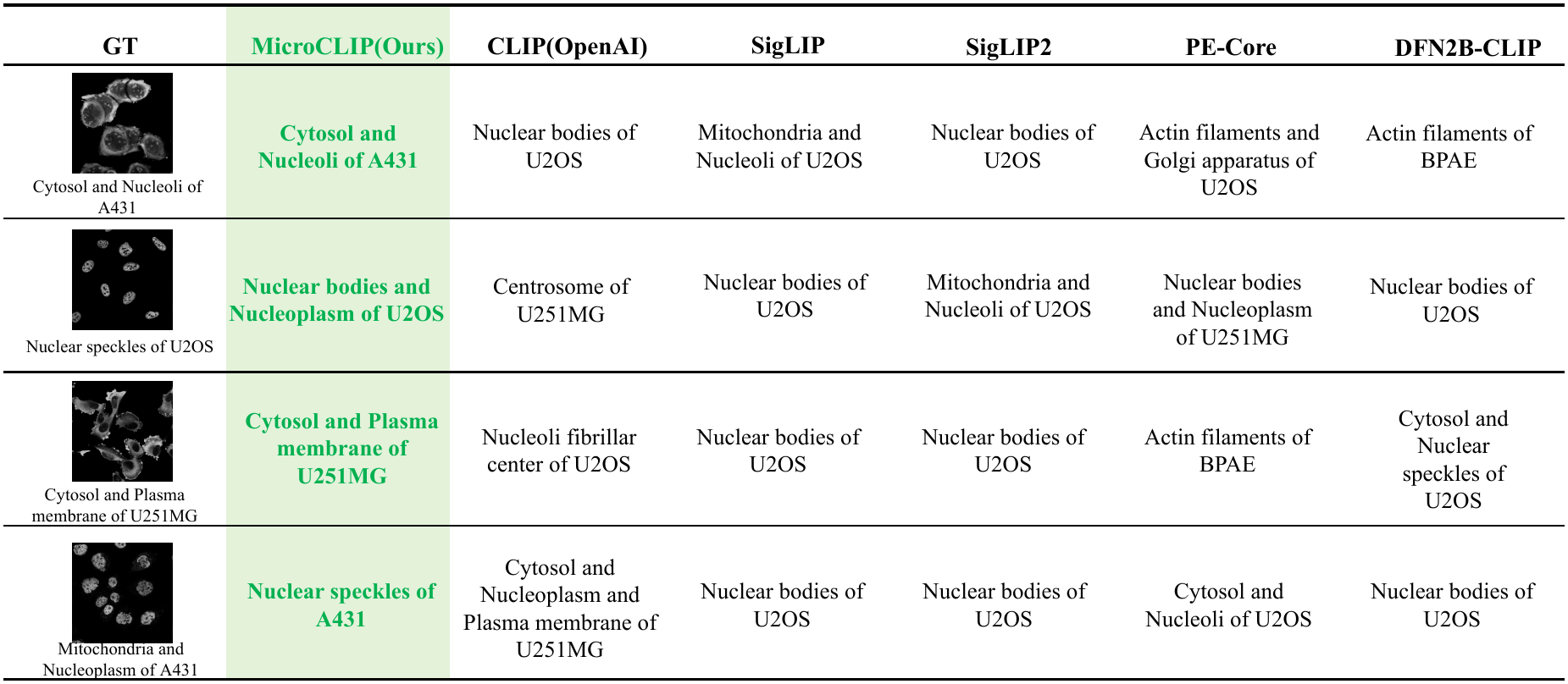}
\caption{Organelle-Set Shift examples comparing model predictions on complete structure labels absent from training, with cell lines represented in training.}
\label{fig:organ}
\end{figure}

\begin{figure}[htbp]
\centering
\includegraphics[width=0.95\textwidth]{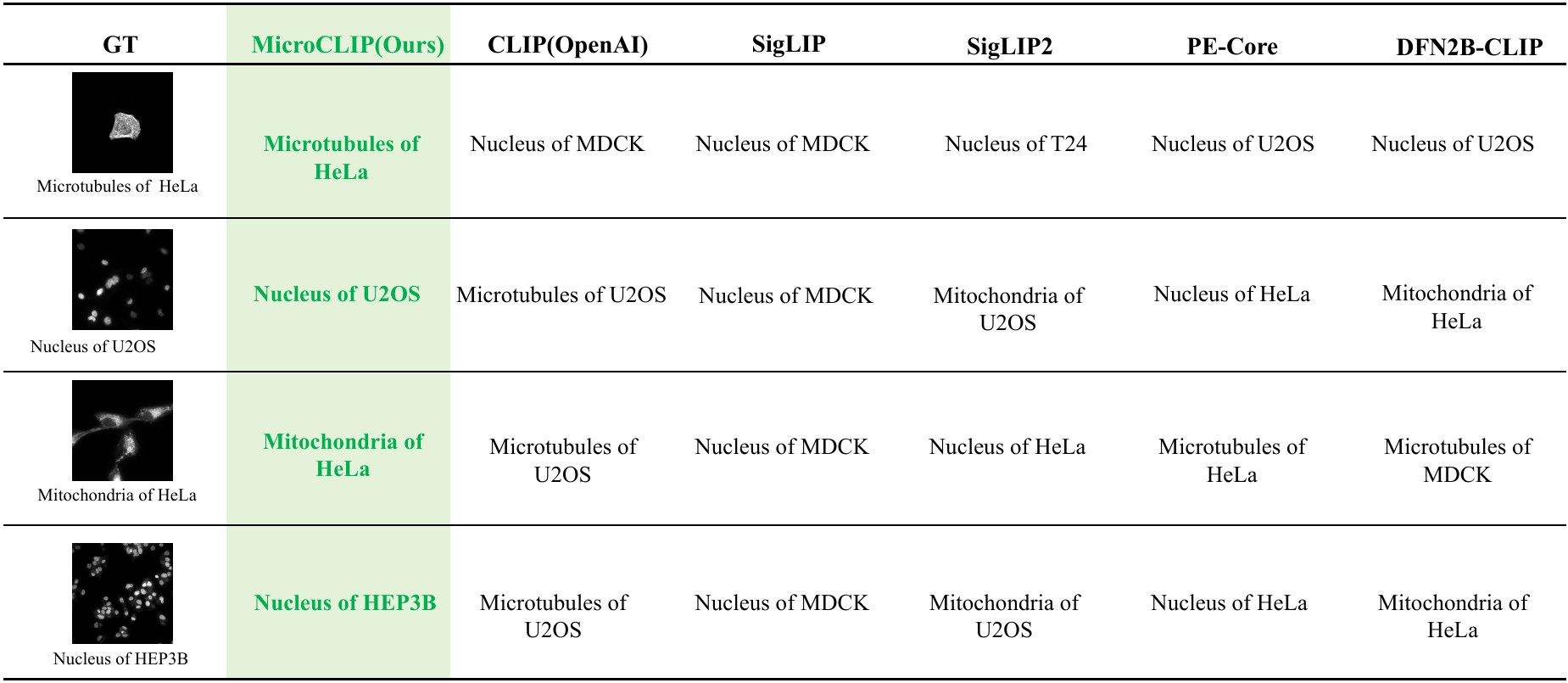}
\caption{Source Shift examples comparing MicroCLIP and baseline predictions with ground-truth labels on images from the external Light My Cells dataset.}
\label{fig:source}
\end{figure}

Figures~\ref{fig:IID}--\ref{fig:source} compare MicroCLIP predictions with ground-truth labels and general-purpose baselines under IID and four distribution shifts. The examples illustrate recognition of subcellular structures and cell lines.

\subsection{Cross-Modal Retrieval}
\label{app:retrieval_visualization}

\begin{figure}[htbp]
\centering
\includegraphics[width=0.95\textwidth]{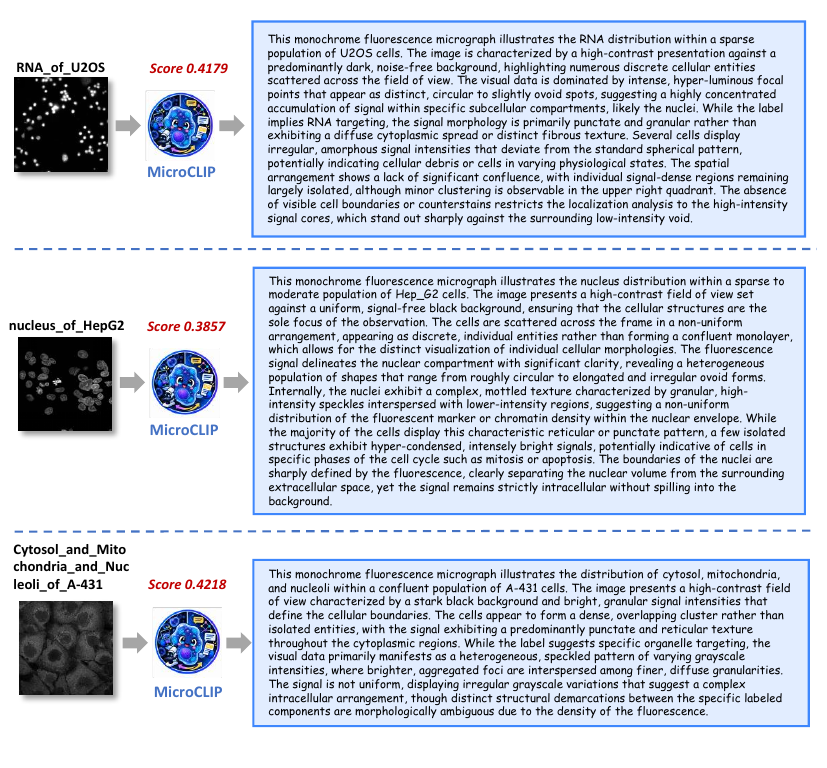}
\caption{Image-to-text retrieval examples: microscopy image queries and retrieved descriptions.}
\label{fig:I2T}
\end{figure}

\begin{figure}[htbp]
\centering
\includegraphics[width=0.95\textwidth]{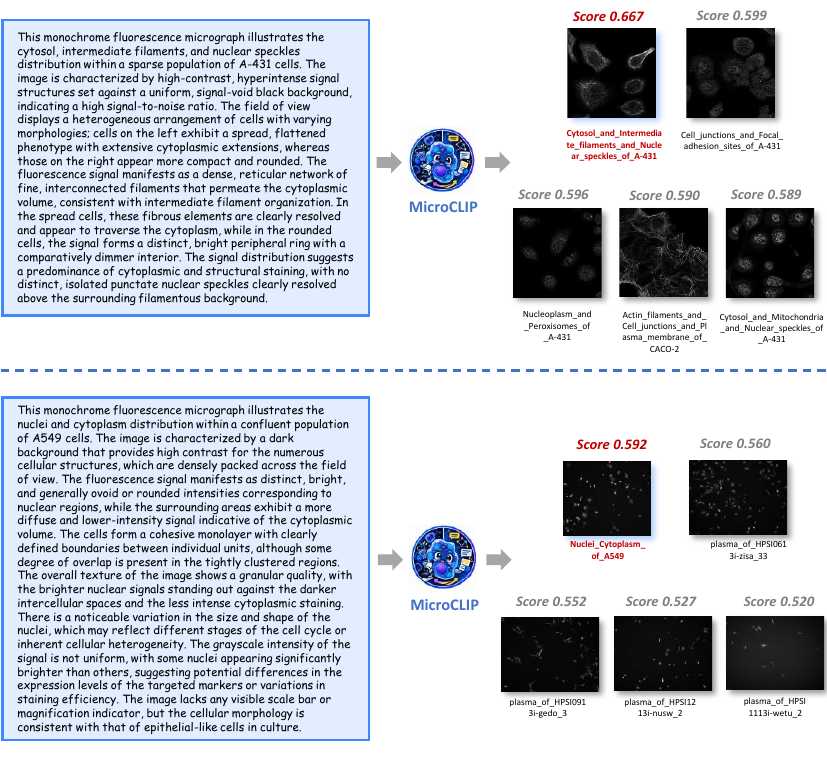}
\caption{Text-to-image retrieval examples: textual queries and retrieved microscopy images.}
\label{fig:T2I}
\end{figure}

Figures~\ref{fig:I2T} and~\ref{fig:T2I} illustrate image-to-text and
text-to-image retrieval, respectively. Given a microscopy image, image-to-text
retrieval ranks candidate descriptions by image--text similarity; in the
reverse direction, a textual description serves as the query for ranking
candidate images. These examples complement the Recall@1 results by showing
the visual and semantic correspondence between queries and retrieved candidates.


\end{document}